\documentclass[final]{siamltex}

\usepackage[dvips]{graphicx}
\usepackage{amsmath}
\usepackage{amssymb}
\usepackage{psfrag}
\usepackage{color}
\usepackage{subfigure}
\usepackage{sidecap}
\newcommand{\bs}{\boldsymbol}
\def\RR{ \mathbb R}

\usepackage{epsf}
\newcommand{\ee}{\end{equation}}
\newcommand{\be}{\begin{equation}}
\newcommand{\ec}{\end{center}}
\newcommand{\bc}{\begin{center}}
\newcommand{\eea}{\end{eqnarray}}
\newcommand{\bea}{\begin{eqnarray}}
\newcommand{\bd}{\begin{description}}
\newcommand{\ed}{\end{description}}
\newcommand{\bi}{\begin{itemize}}
\newcommand{\ei}{\end{itemize}}
\newcommand{\pa}{\partial}
\newcommand{\refeq}[1]{Equation (\ref{#1})}

\newcommand{\pe}{\psi}
 
\def\ds{\displaystyle} 
\def\e{{\epsilon}}

\def\fishpack{{FISHPACK}} 
 
\def\gmres{{GMRES}} 
\def\gmresm{{\rm GMRES($m$)}}

\def\bfE{\mbox{\boldmath$E$}}
\def\bfG{\mbox{\boldmath$G$}}

\title{Scalable Bayesian reduced-order models for simulating high-dimensional multiscale dynamical systems
\thanks{This   work was supported by the OSD/AFOSR MURI'09 award to Cornell University on uncertainty quantification}}

\author{
Phaedon-Stelios Koutsourelakis \thanks{School of Civil and Environmental Enginerring \& Center for Applied Mathematics, 369 Hollister Hall,  Cornell University, Ithaca, NY 14853, ({\tt pk285@cornell.edu})}
 \and Elias Bilionis \thanks{Center for Applied Mathematics, 464 Hollister Hall, Cornell University, Ithaca, NY 14853, ({\tt ib227@cornell.edu}). }
}

\begin{document}

\maketitle

\begin{abstract}
  While existing mathematical descriptions can accurately account for phenomena
at microscopic scales (e.g. molecular dynamics), these are often
high-dimensional, stochastic and their applicability over macroscopic time
scales of physical interest is computationally infeasible or impractical. In
complex systems, with limited physical insight on the coherent behavior of
their constituents, the only available information is data obtained from
simulations of the trajectories of huge numbers of degrees of freedom over
microscopic time scales. This paper discusses a Bayesian approach to deriving
probabilistic coarse-grained models that simultaneously address the problems of
identifying appropriate reduced coordinates and the effective dynamics in this
lower-dimensional representation. At the core of the models proposed lie
simple, low-dimensional dynamical systems which serve as the building blocks of
the global model. These approximate the latent, generating sources and
parameterize the reduced-order dynamics. We discuss parallelizable, online
inference and learning algorithms that employ Sequential Monte Carlo samplers
and scale linearly with the dimensionality of the observed dynamics. We propose
a Bayesian adaptive time-integration scheme that utilizes probabilistic
predictive estimates and enables rigorous concurrent s imulation over
macroscopic time scales. The data-driven perspective advocated assimilates
computational and experimental data and thus can materialize data-model fusion.
It can deal with applications that lack a mathematical description and where
only observational data is available. Furthermore, it makes non-intrusive use
of existing computational models.

\end{abstract}



\pagestyle{myheadings}
\thispagestyle{plain}
\markboth{P.S. KOUTSOURELAKIS, E. BILIONIS}{ Bayesian reduced-order models
}

\section{Introduction}

The present  paper is concerned with the development of probabilistic coarse-grained  models for  high-dimensional dynamical systems with a view of enabling multiscale simulation. We describe a unified treatment of complex problems described by large systems of deterministic or stochastic ODEs and/or large number of data streams.  Such systems arise frequently in modern multi-physics applications either due to the discrete nature of the system (e.g. molecular dynamics) or due to discretization of spatiotemporal models (e.g. PDEs):
\be
\label{eq:master}
\frac{d \bs{y}_t}{dt}=\bs{f}( \bs{y}_t), \quad \bs{y} \in \mathcal{Y}
\ee
where $dim(\mathcal{Y})>>1$ (e.g. $\RR^d$, $d>>1$). Stochastic versions are also frequently encountered: 
\be
\label{eq:master1}
\frac{d \bs{y}_t}{dt}=\bs{f}( \bs{y}_t; \bs{u}_t)
\ee
where $\bs{u}_t$ is a driving stochastic process (i.e. Wiener process). Uncertainties  could also appear in the initial conditions that accompany the aforementioned systems  of equations.

Even though the numerical solution of (stochastic) ODEs is a well-studied subject and pertinent computational libraries are quite mature, traditional schemes are impractical or infeasible  for systems which are high-dimensional and exhibit a large disparity in scales. This is because most numerical integrators must use time-steps of the order of  the fastest scales which precludes solutions over long time ranges that are of interest for physical and engineering purposes.
In the context of atomistic simulations, practically relevant time  scales exceed typical integration steps of $\sim 1fs$ by several orders of magnitude \cite{Abraham:2002}.
Furthermore, when numerical solutions of transient PDEs are sought,  resolution and accuracy  requirements  give rise to systems with more than $10^9$  degrees of freedom \cite{lig08ter,cha08sca,tay08pet,ham08tow} where the integration time steps are slaved by fast reaction rates or high oscillation frequencies. This impedes their solution and frequently constitutes computationally infeasible other important tasks such as stability analysis, sensitivity, design and control.

Multiscale dynamical systems  exist independently of the availability of  mathematical models. Large numbers of time series appear in financial applications, meteorology, remote sensing where the phenomena of interest unfold also over a large range of time scales \cite{wei93tim,Horenko:2007}. A wealth of time series data  is also available in experimental physics and engineering  which by themselves or in combination with mathematical models can be useful in analyzing underlying phenomena \cite{mil99dat,kal02atm,ott04loc} by deriving reduced, predictive descriptions.

Quite frequently the time evolution of all the observables is irrelevant for physical and practical purposes and the analysis is focused on a reduced set of  variables or reaction coordinates  $\hat{\bs{y}}_t=\mathcal{P}(\bs{y}_t)$ obtained by an appropriate mapping $\mathcal{P}:\mathcal{{Y}} \to \mathcal{\hat{Y}}$. The goal is then to identify a closed,   deterministic or stochastic system of equations with respect to 
 $\hat{\bs{y}}_t$, e.g. :         
\be
\label{eq:reduced} 
 \frac{d \hat{\bs{y}}_t}{dt}=\hat{\bs{f}}( \hat{\bs{y}}_t), \quad \hat{\bs{y}}_t \in \mathcal{\hat{Y}}
\ee 
In the context of equilibrium theormodynamics where enmsemble averages with respect to the invariant distribution of $\hat{\bs{y}}_t$ are of interest, coarse-graining amounts to free-energy computations \cite{chi07fre}. In the nonequilibrium case and when an invariant distribution exists, a general approach for deriving effective dynamics is  based on Mori-Zwanzig projections \cite{zwa01non,giv04ext,Chorin:2000,Chorin:2006,Chorin:2005,dar09com}.
 Other powerful numerical approaches to identify the dynamical behavior with respect to the reduced coordinates include  transition path sampling, the transfer operator approach, the nudged elastic band, the string method,  Perron cluster analysis and spectral decompositions  \cite{Dellago:1998,Deuflhard:2000,Deuflhard:2005,E:2005,del06gra, mez05spe}. Marked efforts in chemical kinetics have led to an array of computational tools such as computational singular perturbation \cite{Lam:1993,Lam:1994}, the intrinsic low-dimensional manifold approach \cite{Maas:1992,Zagaris:2004} and others \cite{saw06mod,ren07red,Kavousanakis:2007}. Notable successes in  overcoming the timescale dilemma  have also been achieved in the context of MD simulations \cite{Laio:2002,Voter:2002,Voter:2002a,tuc92rev} (or Hamiltonian systems in general  \cite{pet97num,lei04sim,tao09non}).

In several problems,  physical or mathematical arguments  have led  analysts to identify  a few, salient features and their inter-dependencies that macroscopically describe the behavior of very complex systems consisting of a  huge number of individuals/agents/components/degrees of freedom. These variables parameterize  a low-dimensional, attracting, invariant,  ``slow'' manifold
characterizing the long-term process dynamics \cite{hai02geo}. 
   Hence the apparent  complexity exhibited in the high-dimensionality and the multiscale character  of the original model is a pretext of a much simpler, latent structure that, if revealed, could make the aforementioned analysis tasks much more tractable.   
The emergence of macroscopic, coherent behavior has been the foundation of coarse-grained  dynamic  models that have been  successful  in a wide range of applications. The coarse-grained parameterization and associated model depend on the analysis objectives and particularly on the time scale one wishes  to make predictions.
Modern approaches with general applicability such as the Equation-free method  \cite{Kevrekidis:2004} or Heterogeneous Multiscale Method (HeMM,\cite{e03het})
are also based on the   availability of  reduced coordinates and in the case of HeMM of a macroscopic model which is informed and used in conjunction with the microscale model.

Largely independently of the developments in the fields of computational physics and engineering, the problem of deriving, predictive  reduced-order models for a large number of time series  that potentially exhibit multiple scales has also been addressed in statistics  and machine learning communities \cite{wes97bay,lee07mul} 
with  applications in network analysis \cite{DBLP:conf/sigmod/CranorJSS03}, environmetrics \cite{wik98hie}, sensor network monitoring \cite{DBLP:conf/cidr/YaoG03, DBLP:conf/sigmod/SakuraiPF05}, moving object tracking \cite{DBLP:conf/icde/Aggarwal09}, financial data analysis  \cite{agu00bay}, computer model emulation \cite{liu09dyn}.
Significant advances have been achieved in modeling \cite{DBLP:conf/vldb/TengCY03}, forecasting \cite{DBLP:conf/icde/YiSJJFB00} and developing online, scalable algorithms
 \cite{DBLP:conf/sigmod/FaloutsosKS07,DBLP:conf/pakdd/SunPF06,DBLP:conf/sigmod/KeoghCMP01,DBLP:conf/sigmod/KornJF97,DBLP:journals/cviu/KanthAAS99,DBLP:conf/vldb/ZhuS02}. 
 that are frequently based on the discovery of “hidden variables” that provide insight to the intrinsic structure of streaming data \cite{mul06fer,DBLP:conf/sac/SousaTTF06,DBLP:conf/vldb/PapadimitriouSF05,DBLP:journals/vldb/PapadimitriouBF04,DBLP:conf/vldb/KoudasIM00}.

The present paper proposes a data-driven alternative that is able to automatically coarse-grain high-dimensional systems without the need of preprocessing and availability of physical insight. The data is most commonly obtained by simulations  of the most reliable,  finest-scale (microscopic) model available. This is used to infer  a lower-dimensional description  that captures the dynamic evolution of the system at a  coarser scale (i.e. a macroscopic model). 
The majority of available techniques address separately  the problems of  identifying appropriate reduced coordinates and the effective dynamics in this lower-dimensional representation. It is easily understood that the solution of one affects the other. We propose a general framework where these two problems are  simultaneously solved and  coarse-grained models are built from the ground up. We propose procedures that  concurrently infer the macroscopic dynamics and their mapping  the high-dimensional, fine-scale description. As a result no errors  or ambiguity are introduced when the fine-scale model needs to be reinitialized  in order to obtain additional simulation data. 
To that end, we advocate a largely unexplored in computational physics perspective based on  the Bayesian paradigm which provides a rigorous foundation for learning from data. It is capable of   quantifying inferrential uncertainties and, more importantly, uncertainty due to  information loss in the coarse-graining process.
 
We present a Bayesian state-space model where the reduced, coarse-grained dynamics are parametrized by tractable, low-dimensional  dynamical models. These can be  viewed as experts offering opinions on the evolution of the high-dimensional observables. Each of these modules could represent a single latent regime and would therefore be insufficient by itself to explain the whole system. As is often the case with real experts, their opinions are valid under very specific regimes. We propose therefore a framework for dynamically synthesizing such models  in order to obtain an accurate global representation that retains its interpretability and  computational tractability.

An important contribution of the paper, particularly in view of enabling simulations of multiscale systems, is  online inference algorithms based on  Sequential Monte Carlo that scale linearly with the dimensionality of the observables $d$ (\refeq{eq:master}). These allow the recursive assimilation of data and re-calibration of the coarse-grained dynamics. 
The Bayesian framework adopted  provides probabilistic predictive estimates that can be   employed in the context of adaptive time-integration. Rather than determining integration time steps based on traditional  accuracy and stability metrics, we propose using  measures of the predictive uncertainty in order to decide how long into the future the coarse-grained model can be used.  When the uncertainty associated with the predicitve estimates exceeds the analyst's tolerance, the fine-scale dynamics can be   consistently re-initialized in order to obtain additional data that  sequentially update the coarse-grained  model.

In agreement with some recently proposed methodologies \cite{Kevrekidis:2004,kev03equ}, the  data-driven strategy can seamlessly interact with existing   numerical integrators that are well-understood and reliably implemented in several legacy codes. In addition, it is  suitable for problems where observational/experimental data must be fused with mathematical descriptions 
in a rigorous fashion and lead to improved analysis and prediction tools. 

 The structure of the rest of the paper is as follows. In Section \ref{sec:desc} we describe the proposed framework in relation  with the state-of-the-art in dimensionality reduction. We provide details of the probabilistic model proposed in the context of Bayesian State-Space models  in Section \ref{sec:pmhmm}. Section \ref{sec:inference} is devoted to inference and learning tasks which involve a locally-optimal  Sequential Monte Carlo sampler and an online Expectation-Maximization  scheme. The utilization of the coarse-grained dynamics in the context of a Bayesian (adaptive) time-integration scheme is discussed in \ref{sec:bayesint} and numerical examples are provided in Section \ref{sec:examples}.

\section{Proposed Approach}
\label{sec:desc}


\subsection{From static-linear to dynamic-nonlinear dimensionality reduction}
\label{sec:genintro}

The inherent assumption of all multiscale analysis methods is the existence of a lower-dimensional parameterization of the original system with respect to which the  dynamical evolution is more tractable at the scales of interest.
In some cases these slow variables can be identified a priori  and the problem   reduces to finding the necessary closures that will give rise to a consistent dynamical model.
In general however one must identify the reduced space $\mathcal{\hat{Y}}$ as well as the dynamics within it. 

\begin{figure}
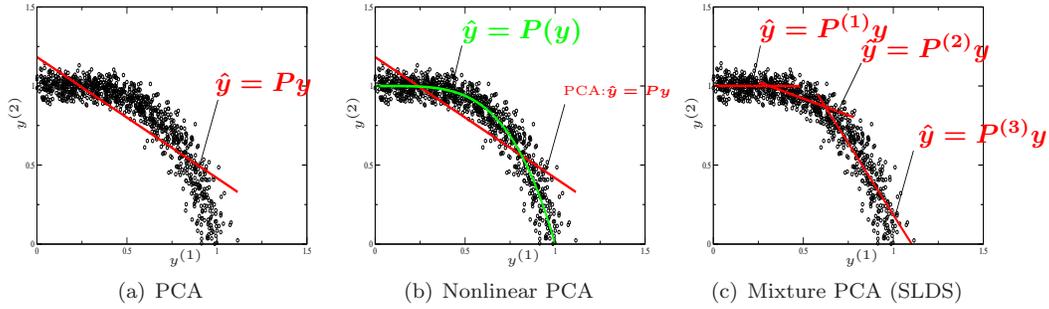

\subfigure[PCA]{
\psfrag{x1}{\tiny  $y^{(1)}$}
\psfrag{x2}{\tiny $y^{(2)}$}
\psfrag{px}{\color{red} $\bs{\hat{y}=Py}$}
\label{fig1a}
\includegraphics[width=0.30\textwidth,height=3.5cm]{FIGURES/pod.eps}} \hfill
\subfigure[Nonlinear PCA]{
\psfrag{x1}{\tiny  $y^{(1)}$}
\psfrag{x2}{\tiny $y^{(2)}$}
 \psfrag{px}{\tiny \color{red} PCA:$\bs{\hat{y}=Py}$}
\psfrag{npx}{\color{green} $\bs{\hat{y}=P(y)}$}
\label{fig1b}
\includegraphics[width=0.30\textwidth,height=3.5cm]{FIGURES/npod.eps}} \hfill
\subfigure[Mixture PCA (SLDS)]{
\psfrag{x1}{\tiny  $y^{(1)}$}
\psfrag{x2}{\tiny $y^{(2)}$}
\psfrag{px1}{\color{red} $\bs{\hat{y}=P^{(1)}y}$}
\psfrag{px2}{\color{red} $\bs{\hat{y}=P^{(2)}y}$}
\psfrag{px3}{\color{red} $\bs{\hat{y}=P^{(3)}y}$}
\label{fig1c}
\includegraphics[width=0.30\textwidth,height=3.5cm]{FIGURES/hmm_pca.eps}} \\
\caption{  \em The phase space is assumed two-dimensional for illustration purposes i.e. $\bs{y}_t=(y^{(1)}_{t}, y^{(2)}_{t})$. Each black circle corresponds to a realization $\bs{y}_{t_i}$. $\mathcal{P}:  \mathcal{Y} \to \mathcal{\hat{Y}} $ is  the projection operator from the original high-dimensional space $\mathcal{Y}$ to the reduced-space $\mathcal{\hat{Y}}$.    }
\end{figure}

A prominent role in these efforts has been held by Principal Component Analysis (PCA) -based methods. With small differences and depending on the community other terms such as Proper Orthogonal Decomposition (POD) or Karhunen-Lo\`eve expansion (KL), Empirical Orthogonal Functions (EOF)  have also been used. PCA finds its roots in the early papers by Pearson \cite{pea01lin} and Hotelling \cite{hot33ana} and was originally developed as a {\em static} dimensionality reduction technique. It is based
on projections on a reduced basis identified by the leading eigenvectors of the covariance matrix $\bs{C}$. In the dynamic case and in the absence of closed form expressions for the actual covariance matrix, samples of the process  $\bs{y}_{t} \in \RR^d$ at $N$ distinct time instants $t_i$ are used in order to  obtain an estimate of the covariance matrix:
\be
\label{eq:cov}
 \bs{C} \approx \bs{C}_N=\frac{1}{N-1} \sum_{i=1}^N (\bs{y}_{t_i} -\bs{\mu}) (\bs{y}_{t_i}-\bs{\mu})^T
\ee
where $\bs{\mu}=\frac{1}{N}  \sum_{i=1}^N \bs{y}_{t_i} $  is the empirical mean. If there is a spectral gap after the first $k$ eigenvalues and  $\bs{V}_K$ is the $d\times K$ matrix whose columns  are the $K$ leading normalized eigenvectors of $\bs{C}_N$ then  the reduced-order model is defined with respect to $\bs{\hat{y}}_t=\bs{V}_K \bs{y}_t$. The reduced dynamics can be identified by a Galerkin projection (or a Petrov-Galerkin projection) of the original ODEs in \refeq{eq:master}:
\be
\label{eq:pca}
\frac {d \bs{\hat{y}}_t}{d t}=\bs{V}^T_K \bs{f}(\bs{V}^T_K \bs{\hat{y}}_t)
\ee
Hence the reduced space $\mathcal{\hat{Y}}$ is approximated by a hyperplane in $\mathcal{Y}$ and the projection mapping $\mathcal{P}$ linear (Figure \ref{fig1a}). While it can be readily shown that the projection adopted is optimal in the mean square sense for stationary Gaussian processes, it is generally not so in cases where non-Gaussian processes  or other distortion metrics are examined. The application of PCA-based techniques, to  high-dimensional, multiscale dynamical systems poses several modeling limitations. Firstly, the reduced  space $\mathcal{\hat{Y}}$  might not be sufficiently approximated by  a hyperplane of dimension $K<<d$. Even though this assumption might hold locally,  it is unlikely that this will be a good global approximation. Alternatively, the dynamics of the original process might be adequately approximated on $K$-dimensional  hyperplane but this hyperplane might change in time. Secondly, despite the fact that the projection on the subspace spanned by the leading eigenvectors captures most of the variance of the original process, in cases where this variability is due to the fast modes, there is no guarantee that $\bs{\hat{y}}_t$ will account for  the long-range, slow  dynamics which is of primary interest in multiscale systems. Thirdly, the basic assumption in the estimation of the covariance matrix, is that the samples $\bs{y}_{t_i}$ are drawn from the same distribution, i.e. that the process $\bs{y}_t$ is  stationary. A lot of multiscale problems however involve  non-stationary dynamics (e.g. non-equilibrium MD \cite{hoo92non,cic05non}).   Hence even if a stationary reduced-order process provides a good, {\em local}, approximation to $\bs{y}_t$, it might need to be updated in time.
Apart from the aforementioned modeling issues, significant computational difficulties are encountered for high-dimensional systems ($d=dim(\mathcal{Y})>>1$) and large datasets ($N>>1$) as the $K$  leading eigenvectors of large matrices  (of dimension proportional to  $d$ or  $N$) need to be evaluated. 
This effort must  be repeated, if more samples become available (i.e. $N$ increases) and an update of the reduced-order model is desirable. Recent efforts have concentrated on developing online versions \cite{war06onl} that circumvent this problem.


The obvious extension to the linear projections of PCA is nonlinear dimensionality reduction techniques. These have been the subject  of intense research in statistics and machine learning in recent years (\cite{sch97ker,Roweis:2000,Tenenbaum:2000,Donoho:2003, Shi:2000,Bach:2006,AzranG06}) and fairly recently have found their way to computational physics and multiscale dynamical systems (e.g. \cite{Coifman:2005,Lafon:2006,Nadler:2006,bas08non}).
%
They are generally based on calculating  eigenvectors of an affinity matrix of a weighted graph. While they circumvent the limiting, linearity assumption of standard PCA, they still assume that the underlying process is stationary (Figure \ref{fig1b}). Even though the system's dynamics might be appropriately tracked on a lower-dimensional subspace for a certain time period, this might not be invariant across the whole time range of
 interest. The identification of the dynamics on the reduced-space $\mathcal{\hat{Y}}$ is not as straightforward as in standard PCA and in most cases, a  deterministic or stochastic  model is fit directly to the projected data points  \cite{Coifman:2008,Erban:2007,fra08hid}. More importantly since the inverse mapping $\mathcal{P}^{-1}$ from the manifold  $\mathcal{\hat{Y}}$ to $\mathcal{Y}$ is not available analytically, approximations have to be made in order to find pre-images in the data-space \cite{DBLP:conf/dagm/BakirZT04,Erban:2007}.  From a computational point of view, the cost of identifying the projection mapping is comparable to standard PCA as an eigenvalue problem on a $N \times N$ matrix has to be solved. Updating those eigenvalues and the nonlinear projection operator in cases where additional data become available  implies a significant computational overhead although recent efforts \cite{sch07fas} attempt to overcome this limitation.


A common characteristic of the aforementioned techniques is that even though the reduced coordinates are learned from {\em a finite amount of simulation data}, there is no {\em quantification of the uncertainty} associated with these inferences. This is a critical component not only in cases where multiple sets of reduced parameters and coarse-grained models are consistent with the data, but also for assessing errors associated with the analysis and prediction estimates. It is one of the main motivations for adopting a {\em probabilistic approach} in this project. Statistical models can naturally deal with stochastic systems that frequently arise in a lot of applications. Most importantly perhaps, even in cases where the fine-scale model is deterministic (e.g. \refeq{eq:master}),  a stochastic reduced model provides a better  approximation that can simultaneously quantify the uncertainty arising from the information loss that takes place during the coarse-graining process \cite{fat04com,kou07sto}. 


%

%


A more general perspective is offered by latent variable models where  the observed data (experimental or computationally generated) is augmented by a set of hidden variables  \cite{bis99lat}. {\em In the case of high-dimensional, multiscale dynamical systems, the latent model corresponds to a reduced-order process  that evolves at scales of practical relevance.} Complex distributions over the observables can be  expressed in terms of simpler and tractable joint distributions over the expanded variable space.  Furthermore, {\em structural characteristics} of the  original, high-dimensional  process $\bs{y}_t$  can be revealed by interpreting the latent variables  as generators of the observables. 


In that respect, a general setting is  offered by Hidden Markov Models (HMM, \cite{gha01int}) or more generally State-Space Models (SSM) \cite{cap01ten,gha04uns,Horenko:2007}. These assume the existence of an {\em unobserved (latent)} process $\bs{\hat{y}}_t \in \RR^K$ described by a (stochastic) ODE:

\be
\label{eq:ssm1}
\frac{d \bs{\hat{y}}_t}{dt}=\bs{\hat{f}}(\bs{\hat{y}}_t; \bs{w}_t) \quad (\textrm{transition equation})
\ee
which gives rise to the observables $\bs{y}_t \in \RR^d$ as: 

\be
\label{eq:ssm2}
\bs{y}_t=\bs{h}(\bs{\hat{y}}_t, \bs{v}_t) \quad (\textrm{emission equation})
\ee
where $\bs{w}_t$ and $\bs{v}_t$ are unknown stochastic processes (to be inferred from data)  and $\bs{\hat{f}}: \RR^K \to \RR^K$, $\bs{h}: \RR^K \to \RR^d$ are unknown measurable  functions. The transition equation defines a prior distribution on the coarse-grained dynamics whereas the emission equation, the mapping  that connects the reduced-order representation with the observable dynamics. The object of Bayesian inference is to learn the unobserved (unknown) model parameters from the observed data. Hence the coarse-grained model and its relation to the observable dynamics are inferred from the data. 

The form of Equations (\ref{eq:ssm1}) and (\ref{eq:ssm2})  affords  general representations. Linear and nonlinear PCA models arise as special cases by appropriate selection of the functions and random processes appearing in the transition and emission equations. Note for example that the  transition equation (\refeq{eq:ssm1}) for $\bs{\hat{y}}_t$ in the case of the PCA-based models reviewed earlier is given by \refeq{eq:pca} and the {\em emission equation} (\refeq{eq:ssm2}) that relates latent and observed processes is linear, deterministic  and specified by the matrix of $K$ leading eigenvectors $\bs{V}_K$. 

 An extension to HMM is offerered by switching-state models \cite{har76bay,cha78sta,ham89new,shu91dyn}  which can be thought of as dynamical mixture models \cite{DBLP:journals/tsmc/ChaerBG97,DBLP:journals/neco/GhahramaniH00}. The latent dynamics  consist of a discrete process that  takes  $M$ values, each corresponding to a distinct dynamical behavior. This can be represented by  an $M$-dimensional vector   $\bs{z}_t$  whose  entries are zero except for a single one $m$ which is equal to one and represents the active mode/cluster. Most commonly, the time-evolution of $\bs{z}_t$ is modeled by a first-order stationary Markov process:
\be
\label{eq:slds}
\bs{z}_{t+1}=\bs{T} \bs{z}_t
\ee
where $\bs{T}=[T_{m,n}]$ is the transition matrix and $T_{m,n}=Pr[z_{m,t+1}=1 \mid z_{n,t}=1]$.
 In addition to $\bs{z}_t$, 
 $M$ processes $\bs{x}_t^{(m)} \in \RR^K, ~m=1,\ldots,M$ parameterize the reduced-order  dynamics (see also discussion in section \ref{sec:pmhmm}).  Each is activated when $z_{m,t}=1$. In the linear version (Switching Linear Dynamic System, SLDS \footnote{sometimes referred to as jump-linear or conditional Gaussian models}) and conditioned on $z_{m,t}=1$, the observables $\bs{y}_t$ arise by a projection from the active $\bs{x}_t^{(m)}$ as follows:
\be
\label{eq:obs_slds}
\bs{y}_t= \bs{P}^{(m)} \bs{x}_{t}^{(m)}+\bs{v}_t, \quad \bs{v}_t \sim N(\bs{0}, \bs{\Sigma})~(i.i.d)
\ee
where  $\bs{P}^{(m)}$ are $d\times K$  matrices ($K<<d$) and $\bs{\Sigma}$ is a positive definite $d\times d$ matrix.
Such models provide a natural, physical interpretation according to which the behavior of the original process $\bs{y}_t$ is segmented into  $M$ regimes or clusters, the dynamics of which can be low-dimensional and tractable. From a modeling point of view, the idea of utilizing a mixture of simple models provides great flexibility \cite{ble03lat,ble03hie,sre05tim,gri06inf} as it can be theorized that given a  large enough number of such components, any type of dynamics can be sufficiently approximated. In practice however, a large  number  might be needed, resulting in complex models containing a large number of parameters. 

Such mixture models have gained prominence in recent years in the machine learning community. In \cite{ble06dyn} for example, a dynamic  mixture model was used to analyze a huge number of time series, each corresponding to a word in the English vocabulary as they appear in  papers in the journal {\em Science}. The latent discrete variables represented {\em topics} and each topic implied a distribution on the space of words. As a result,  not only a predictive summary (dimensionality reduction) of the high-dimensional observables was achieved  but also an insightful deconstruction of the original time series was made possible. In fact current research in statistics has focused on {\em infinite } mixture models where the number of components can be automatically inferred from the data  (\cite{teh06hie,car07gen,bea02inf,DBLP:conf/icml/FoxSJW08,DBLP:conf/nips/FoxSJW08}).
In the context of computer simulations of high-dimensional systems, such models have been employed by \cite{Fischer:2007,Horenko:2006,Horenko:2007,Horenko:2008,hor07dat}
where  maximum likelihood techniques were used to learn the model parameters.

In the next sections we present a novel model that generalizes SLDS. Unlike mixture models which assume that  $\bs{y}_t$ is  the result of a single reduced-order process at a time,  we propose a {\em partial-membership} model (referred to henceforth as Partial-Membership Linear Dynamic System, PMLDS)  which allows observables to have fractional memberships in multiple clusters.  The latent, building blocks are  {\em experts } \cite{jac91ada,jac94hie,hin02tra,hel07non} which, on their own, provide an incomplete, biased prediction but when their ``opinions'' are appropriately synthesized,  they can give rise to a highly accurate aggregate model.

From a modeling perspective such an approach has several appealing properties. The integrated coarse-grained model can be interpretable and low-dimensional even for large, multiscale systems as its expressive ability does not hinge upon the complexity of the individual components but rather is a result of its  factorial  character (\cite{gha99fac}). Intricate dynamical behavior can be captured and decomposed in terms of simple building blocks. It is highly-suited for problems  that lack {\em scale separation} and where the evolution of the system is the result of  phenomena at a {\em cascade of scales}. Each of these scales can be described by a latent  process and the resulting coarse-grained model   will account not only for the  {\em slow} dynamics but also quantify the predictive uncertainty due to the condensed, fast-varying features.

From an algorithmic point of view we present  {\em parallelizable, online} inference/learning  schemes, which can  recursively update the estimates  produced  as more data become available i.e. if the time horizon $t$ of the observables $\bs{y}_{1:t}$ increases. Unlike some statistical applications where long time series are readily available,  in the majority of problems involving computational simulations  of high-dimensional, multiscale systems, data is expensive (at least over large time horizons)  as they imply calls to the microscopic  solvers. The algorithms presented are capable of   producing predictive estimates ``on the fly'' and if additional data is incorporated, they can readily update the model parameters.  In addition, such schemes can take advantage of  the natural  tempering effect of introducing the data sequentially which can further  facilitate the solution of the global estimation problem.  More importantly perhaps, the updating schemes discussed  have {\em linear complexity} with respect to the dimensionality $d$ of the original process $\bs{y}_t$. 

%



\subsection{Partial-Membership Linear Dynamic System}
\label{sec:pmhmm}
We present a hierarchical Bayesian framework which promotes sparsity, interpretability and efficiency. 
The framework described  can integrate heterogeneous building blocks and allows for  physical insight  to be introduced on a case-by-case basis. 
When dealing with high-dimensional molecular ensembles for example, each of these building blocks might be an (overdamped) Langevin equation with a harmonic potential \cite{can07the,Horenko:2007,Horenko:2008b}. It is obvious that such a simplified model would perhaps provide a good approximation under specific, limiting conditions (e.g. at a persistent metastable state) but definitely not across the whole time range of interest. Due to its simple parameterization and computational tractability, it can easily be trained to represent one of the ``experts'' in the integrated reduced-model.  
It is known that these models work well under specific regimes but none of them gives an accurate global description. In the framework proposed, they can however be utilized in a way that combines their strengths, but also probabilistically quantifies their limitations.

A transient, nonlinear PDE can be resolved into several linear PDEs whose simpler form and parameterization makes them computational tractable over macroscopic time scales and permits a  coarser spatial discretization. Their combination with time-varying characteristics can give rise to an accurate global approximation. 
 Their simplicity and limited range of applicability would preclude their individual use. In the framework proposed however,  these simple models would only serve as good local approximants and their inaccurate predictions would be synthesized into an accurate, global model.

\begin{SCfigure}
\psfrag{hid}{latent processes}
\psfrag{time step}{time step}
\psfrag{fast}{\color{red} fast $x_t^{(1)}$}
\psfrag{slow}{\color{blue} slow $x_t^{(2)}$}
\includegraphics[width=0.45\textwidth,height=3.5cm]{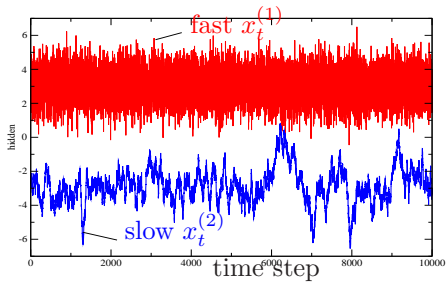}
\label{fig:hid}
\caption{\em Realizations of two hidden ($M=2$) one-dimensional ($K=1$) Ornstein-Uhlenbeck processes were used $d\bs{x}^{(m)}_t=-b^{(m)} (\bs{x}_t^{(m)}-\bs{q}_x^{(m)})dt+(\bs{\Sigma}^{(m)})^{1/2} d\bs{W}_t$ with $(b^{(1)},\bs{q}_x^{(1)},\bs{\Sigma}^{(1)})=(1.,3.,2.)$ (fast) and $(b^{(2)},\bs{q}_x^{(2)},\bs{\Sigma}^{(2)})=(0.01,-3.,0.02)$ (slow)} 

\end{SCfigure}


\subsubsection{Representations with simple building blocks}
\label{repres}
We describe a probabilistic, dynamic, continuous-time, generative model which relates  a sequence the observations  $\bs{y}_t \in \RR^d$ at discrete time instants $t=1,2,\ldots \tau$   with  a number of hidden processes. 
The proposed model consists of $M$ hidden processes $\bs{x}_t^{(m)} \in \RR^K,~m=1,\ldots, M $ ($K<<d$) which are assumed to evolve independently of each other and are described by a set of (stochastic) ODEs:
\be
\label{eq:hidx}
\frac{d \bs{x}_t^{(m)} }{dt}=\bs{g}_m(\bs{x}_t^{(m)} ; \bs{\theta}_x^{(m)}), \quad m=1,\ldots,M
\ee
This equation  essentially implies a {\em prior distribution} on the space of hidden processes parameterized by a set of (unknown a priori) parameters $\bs{\theta}_x^{(m)}$. 
It should be noted that while the proposed framework allows for any type of process in \refeq{eq:hidx}, it is desirable that  these are simple, in the sense that the parameters  $\bs{\theta}_x^{(m)}$ are low-dimensional and  can be learned swiftly and efficiently.  We list some desirable properties of the prior models \cite{sre05tim}:
\bi
\item    Stationarity: Unless specific prior information is available, it would be unreasonable to impose a time bias on the evolution of any of the reduced dynamics processes.  Hence it is important that the models adopted are a priori stationary. Note that the posterior distributions might still exhibit non-stationarity.

\item Correlation Decay:  It is easily understood that for any $m$ and $t_1,t_2$, the correlation $\bs{x}_{t_1}^{(m)}$ and $\bs{x}_{t_2}^{(m)}$ should decay monotonically as $\mid t_2-t_1\mid$ goes to $+\infty$. This precludes models that do not explicitly account for the time evolution of the latent processes and assume that hidden states are not time-dependent (e.g. static PCA models). 
\item Other: Although this is not necessary,  we adopt a continuous time model in the sense of \cite{wan08con} with an analytically available transition density which allows inference to be carried out seamlessly even in cases where the observables are  obtained  at non-equidistant times. As a result the proposed framework can adapt to the granularity of the observables and also provide exact probabilistic   predictions at any time resolution.

\ei

\begin{figure}[!t]
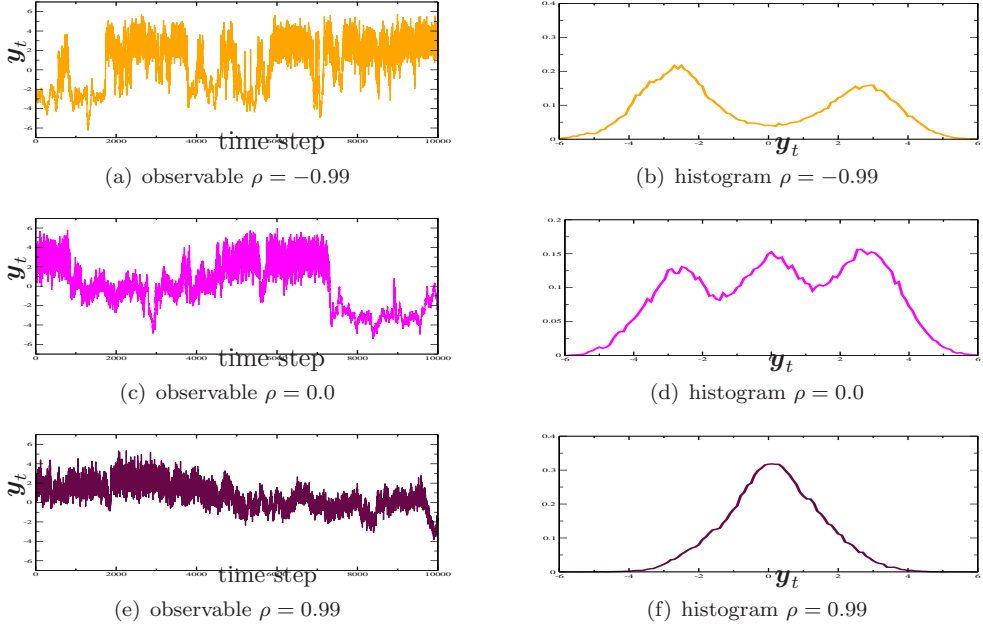

\subfigure[observable $\rho=-0.99$]{
\label{fig:obs1}
\psfrag{time step}{time step}
\psfrag{obs}{ $\bs{y}_t$ }
\includegraphics[width=0.45\textwidth,height=2.cm]{FIGURES/doe_proposal_obs_negative_cor.eps}} \hfill
\subfigure[histogram $\rho=-0.99$]{
\label{fig:hist1}
\psfrag{y}{ $\bs{y}_t$ }
\includegraphics[width=0.45\textwidth,height=2.cm]{FIGURES/doe_proposal_obs_hist_negative_cor.eps}}   \\
\subfigure[observable $\rho=0.0$]{
\label{fig:obs2}
\psfrag{time step}{time step}
\psfrag{obs}{ $\bs{y}_t$ }
\includegraphics[width=0.45\textwidth,height=2.cm]{FIGURES/doe_proposal_obs_zero_cor.eps}} \hfill
\subfigure[histogram $\rho=0.0$]{
\label{fig:hist2}
\psfrag{y}{$\bs{y}_t$ }
\includegraphics[width=0.45\textwidth,height=2.cm]{FIGURES/doe_proposal_obs_hist_zero_cor.eps}} \\
\subfigure[observable $\rho=0.99$]{
\label{fig:obs3}
\psfrag{time step}{\small time step}
\psfrag{obs}{ $\bs{y}_t$ }
\includegraphics[width=0.45\textwidth,height=2.cm]{FIGURES/doe_proposal_obs_positive_cor.eps}} \hfill
\subfigure[histogram $\rho=0.99$]{
\label{fig:hist3}
\psfrag{y}{ $\bs{y}_t$ }
\includegraphics[width=0.45\textwidth,height=2.cm]{FIGURES/doe_proposal_obs_hist_positive_cor.eps}} \\
\caption{\em 
The logistic normal distribution was used to model the weights associated with each of the hidden processes depicted in Figure \ref{fig:hid}. In particular, an  isotropic Ornstein-Uhlenbeck process $d\mathbf{z}_t=-b_z (\mathbf{z}_t-\bs{q}_z)dt+\bs{\Sigma}_z^{1/2} d\bs{W}_t$ with $b_z=0.001$, $\bs{q}_z=[0,~0]^T$ and $\bs{\Sigma}_z$.
 Graphs depict the resulting observable time history (left column) and its histogram (right column) arising from the unobserved processes in Figure \ref{fig:hid} and for three values of $\rho$. It is noted that time histories exhibit fast and slow scales of the processes in Figure \ref{fig:hid}. Furthermore, by  changing a single parameter (i.e. $\rho$)  one can obtain two, three or a single {\em metastable} state (right column - peaks of the histogram).
}
\label{fig:obs}
\end{figure}

Although more complex models can be adopted we assume here that independent,  isotropic Ornstein-Uhlenbeck (OU) processes are used to model the hidden dynamics $\bs{x}_t^{(m)}$. The OU processes used comply with  the aforementioned desiderata. In particular, the following parameterization is employed:
\be
\label{eq:oux}
d \bs{x}_t^{(m)}=-b_x^{(m)} (\bs{x}_t-\bs{q}_x^{(m)})dt+(\bs{S}_{x}^{(m)})^{1/2}d\bs{W}_t^{(m)}
\ee
where $\bs{W}_t^{(m)}$ are Wiener processes (independent for each $m$), $b_x^{(m)}>0$, $\bs{q}_x^{(m)} \in \RR^K$ and  $\bs{S}_{x}^{(m)}$ are positive definite matrices of dimension $K\times K$.
The aforementioned model has a Gaussian invariant (stationary) distribution $\mathcal{N}(\bs{q}_x^{(m)}, \frac{1}{2b_x^{(m)} }\bs{S}_{x}^{(m)})$. The transition density denoted by $p(\bs{x}_t^{(m)} \mid \bs{x}_{t-1}^{(m)})$ for time separation $\delta t$ is also a Gaussian $\mathcal{N}(\bs{\mu}_{t,\delta t}, \bs{S}_{\delta t})$ where:
\be
\label{eq:outrans}
\begin{array}{l}
 \bs{\mu}_{t,\delta t}=\bs{x}_{t-1}^{(m)}-(1-e^{-b_x^{(m)} \delta t})(\bs{x}_{t-1}^{(m)}-\bs{q}_x^{(m)}) \\
\bs{S}_{\delta t} =\frac{1-e^{-2b_x^{(m)} \delta t} }{2 b_x^{(m)} } \bs{S}_{x}^{(m)}
\end{array}
\ee

\begin{figure}
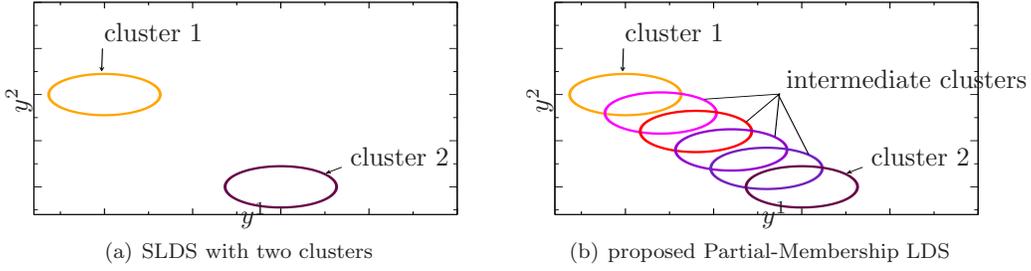

\subfigure[SLDS with two clusters]{
\psfrag{y1}{$y^{1}$}
\psfrag{y2}{$y^{2}$}
\psfrag{c1}{cluster 1}
\psfrag{c2}{cluster 2}
\includegraphics[width=0.45\textwidth,height=3.cm]{FIGURES/mm.eps}
}
\hfill %
\subfigure[proposed Partial-Membership LDS]{
\psfrag{y1}{$y^{1}$}
\psfrag{y2}{$y^{2}$}
\psfrag{c1}{cluster 1}
\psfrag{c2}{cluster 2}
\psfrag{inter}{intermediate clusters}
\includegraphics[width=0.45\textwidth,height=3.cm]{FIGURES/mm1.eps}
}
\caption{Comparison of SLDS  (a) with two mixture components and the proposed model partial-membership model (b) }
\label{fig:comparison}

\end{figure}

It is not expected that simple processes on their own will provide good approximations to the essential dynamics exhibited in the data $\bs{y}_t$.  
In order to combine the  dynamics implied by the $M$ processes in \refeq{eq:hidx}, we consider an $M-$dimensional process $\mathbf{z}_t$ such that $\sum_{m=1}^M z_{m,t}=1$ and $z_{m,t}>0, ~\forall t$ and define an appropriate prior.
The coefficients $z_{m,t}$ express the {\em weight} or {\em fractional membership}  to  each process/expert $\bs{x}_t^{(m)}$ at time $t$ \cite{hel08sta}.  We use the {\em logistic-normal} model \cite{ble06dyn} as a prior for $\mathbf{z}_t$. It is based on a Gaussian process, $\bs{\hat{z}}_t$ whose dynamics are also prescribed by an isotropic OU process:
\be
\label{eq:ouz}
d \bs{\hat{z}}_t=-b_{z} (\bs{\hat{z}}_t-\bs{q}_{z})dt+\bs{S}_{z}^{1/2}d\bs{W}_t
\ee
 and the transformation:
\be
\label{eq:ln}
z_{m,t}=\frac{ e^{\hat{z}_{m,t}} +1/M}{ \sum_{m=1}^M e^{\hat{z}_{m,t}}+1 }, ~ \forall m, t
\ee
The invariant and transition densities of $\bs{\hat{z}}_t$ are obviously identical to the ones for $\bs{x}_t^{(m)}$ with appropriate substitution of the parameters. The hidden processes $\{\bs{x}_t^{(m)}\}_{m=1}^M$ and associated weights $\mathbf{z}_t$   give rise to the observables $\bs{y}_t$  as follows (compare with \refeq{eq:obs_slds}):
\be
\label{eq:obs}
\bs{y}_t=\sum_{m=1}^M z_{m,t} ~\bs{P}^{(m)} \bs{x}_{t}^{(m)}+\bs{v}_t, \quad \bs{v}_t \sim N(\bs{0}, \bs{\Sigma})~(i.i.d)
\ee
where $\bs{P}^{(m)}$ are $d\times K$  matrices ($K<<d$) and $\bs{\Sigma}$ is a positive definite $d\times d$ matrix.  The aforementioned equation implies a series of linear projections  on hyperplanes of dimension $K$. The dynamics along those hyperplanes are dictated by a priori independent process $\bs{x}_t^{(m)}$. It is noted however the reduced dynamics are simultaneously described by {\em all} the hidden processes  (Figure \ref{fig:comparison}). This is in  contrast to PCA   methods where a single  such projection is considered and   mixture PCA models where even though several hidden processes are used, at each time instant it is assumed that a single one is active.  Due to the factorial nature of the proposed model, multiple dynamic regimes can be captured by appropriately combining a few latent states. While mixture models (Figure \ref{fig1c}) provide a flexible framework, the number of hidden states might be impractically large. As it is pointed out in \cite{gha99fac}, in order to encode for example a time sequence with $30 bits$ of information one would need $k=2^{30}$ distinct states. It is noted that even though  {\em linear } projections are implied by \refeq{eq:obs} and Gaussian noise $\bs{v}_t$ is used, the resulting model for $\bs{y}_t$ is {\em nonlinear} and {\em non-Gaussian}  as it involves the factorial combination of $M$ processes $\{\bs{x}_t^{(m)}\}_{m=1}^M$ with  $\mathbf{z}_t$ which are {\em a posteriori} non-Gaussian. 

The parameters $z_{m,t}$ express the relative  importance of the various reduced models or equivalently  the degree to which each data point $\bs{y}_t$ is associated with each of the $M$ reduced dynamics $\bs{x}_t^{(m)}$. It is important to note that the proposed model allows for time varying weights $\bs{z}_{m,t}$ and can therefore account for the possibility of  switching between different regimes of  dynamics.  Figure $\ref{fig:obs}$ depicts a simple example ($d=1$) which illustrates the flexibility of the proposed approach.

The unknown  parameters of the coarse-grained model consist of:
\bi
\item  {\em dynamic} variables denoted for notational economy by $\bs{\Theta}_t$  (i.e. $\{\bs{x}_t^{(m)}\}_{m=1}^M$, $\mathbf{z}_t$ , for  $t=1,2,\ldots $).
\item {\em static} variables denoted by $\bs{\Theta}$ (i.e. $\bs{\theta}_x^{(m)}=(b_x^{(m)}, \bs{q}_x^{(m)}, \bs{S}_{x}^{(m)})$ in \refeq{eq:oux}, $\bs{\theta}_z=(b_{z}, \bs{q}_{z}, \bs{S}_{z})$ in \refeq{eq:ouz} and $\{\bs{P}^{(m)} \}_{m=1}^M, \bs{\Sigma}$ in \refeq{eq:obs}). 
\ei

%

\subsection{Inference and learning}
\label{sec:inference}
Inference in the probabilistic graphical model described involves determining the probability distributions associated with the unobserved (hidden) static 
$\bs{\Theta}$ and dynamic parameters $\bs{\Theta}_t$ of the model.  
In the Bayesian setting adopted this is the {\em posterior } distribution of the  unknown  parameters of the coarse-grained model. 
Given the observations (computational or experimental) of the original, high-dimensional process $\bs{y}_{1:\tau}=\{ \bs{y}_t \}_{t=1}^\tau$, we denote the posterior  by $\pi(\bs{\Theta}, \bs{\Theta}_{1:\tau} )$:
\be
\label{eq:post}
\pi(\bs{\Theta}, \bs{\Theta}_{1:\tau} )= p(\bs{\Theta}, \bs{\Theta}_{1:\tau} \mid \bs{y}_{1:\tau}) = \frac{ \overbrace{ p(\bs{y}_{1:\tau} \mid \bs{\Theta}, \bs{\Theta}_{1:\tau}) }^{likelihood}~\overbrace{ p(\bs{\Theta}_t, \bs{\Theta}) }^{prior} }{ p(\bs{y}_{1:\tau})}
\ee
The normalization constant $  p(\bs{y}_{1:\tau})$ is not of interest when sampling for $\bs{\Theta}$ or $\bs{\Theta}_{1:\tau}$ but can be quite useful for model validation purposes.

A telescopic decomposition holds for the likelihood which  according to \refeq{eq:obs} is given by:
\be
\label{eq:like}
p( \bs{y}_{1:\tau} \mid \bs{\Theta}, \bs{\Theta}_{1:\tau}) = \prod_{t=1}^{\tau}  p( \bs{y}_{t} \mid \bs{\Theta}, \bs{\Theta}_{t})
\ee
where the densities in the product are described in \refeq{eq:likec}. \refeq{eq:obs} defines the  likelihood $p( \bs{y}_{t} \mid \bs{\Theta}, \bs{\Theta}_{t})$ which is basically the {\em weighted product} of the likelihoods under each of the hidden processes/experts $\bs{x}_t^{(m)}$ :
\be
\label{eq:like1}
p( \bs{y}_{t} \mid \bs{\Theta}, \bs{\Theta}_{t})=\frac{1}{c(\bs{\Theta}, \bs{\Theta}_{t}) } \prod_{m=1}^M p_m^{z_{m,t} }(\bs{y}_{t} \mid  \bs{\Theta}, \bs{\Theta}_{t})
\ee
where the normalizing constant  $c(\bs{\Theta}, \bs{\Theta}_{t})$ ensures that the density intergrates to one with respect to $\bs{y}_t$. According to \refeq{eq:obs}:
\be
\label{eq:likem}
p_m(\bs{y}_{t} \mid  \bs{\Theta}, \bs{\Theta}_{t}) \propto \frac{1}{\mid \bs{\Sigma} \mid^{1/2} } \exp \left\{-\frac{1}{2} (\bs{y}_t - \bs{P}^{(m)} \bs{x}_{t}^{(m)})^T \bs{\Sigma}^{-1} (\bs{y}_t - \bs{P}^{(m)} \bs{x}_{t}^{(m)}) \right\}
\ee
The likelihood can be written in a more compact form as:
\be
\label{eq:likec}
p( \bs{y}_{t} \mid \bs{\Theta}, \bs{\Theta}_{t}) \propto \frac{1}{\mid \bs{\Sigma} \mid^{1/2} }  \exp \left\{-\frac{1}{2} (\bs{y}_t - \bs{W}_t \bs{X}_{t})^T \bs{\Sigma}^{-1} (\bs{y}_t - \bs{W}_t \bs{X}_{t}) \right\}
\ee
where:
\be
\label{eq:xtot}
\underbrace{\bs{X}_t^T}_{MK\times 1}=\left[ (\bs{x}_{t}^{(1)})^T,(\bs{x}_{t}^{(2)})^T, \ldots (\bs{x}_{t}^{(M)})^T \right]^T
\ee
and:
\be
\underbrace{ \bs{W}_t}_{d \times MK}=\left[ \begin{array}{lllllll}
z_{1,t} & \bs{P}^{(1)} &  z_{2,t} & \bs{P}^{(2)} & \ldots &  z_{M,t} & \bs{P}^{(M)}\\
         \end{array} \right]
\ee

The first-order Markovian processes  adopted for the prior modeling of the dynamic parameters $\bs{\Theta}_t$ (Equations (\ref{eq:oux}), (\ref{eq:ouz}), \ref{eq:outrans}) ) imply that:
\be
\label{eq:priortot}
p(\bs{\Theta}_{1:\tau}, \bs{\Theta})=p(\bs{\Theta}) \prod_{t=1}^{\tau} p(\bs{\Theta}_t \mid \bs{\Theta}_{t-1}, \bs{\Theta}) 
\ee
where $p(\bs{\Theta}_1 \mid \bs{\Theta}_{0}, \bs{\Theta})=p(\bs{\Theta}_1 \mid  \bs{\Theta})=\nu_{0}(\bs{\Theta}_1  \mid \bs{\Theta})$ is the prior on the initial condition which in this work is taken to be the stationary distribution of the underlying  OU processes (see discussion in Section \ref{repres}) and denoted for notational economy by $\nu_{0}(. \mid \bs{\Theta})$.

  The posterior   encapsulates uncertainties arising from the potentially stochastic nature of the original process $\bs{y}_t$ as well as due to the fact that a finite number of observations were used.  
The difficulty of the problem is that both the   dynamic ($\bs{\Theta}_{1:\tau}$) and the static parameters ($\bs{\Theta}$) are unknown. 
We adopt a hybrid strategy whereby we sample from the full posterior for the dynamic parameters $\bs{\Theta}_t$ and provide point estimates for the static parameters $\bs{\Theta}$. If uniform priors are used for $\bs{\Theta}$ then the procedure proposed reduces to a maximum likelihood estimation. Non-uniform priors have a regularization effect which can promote the identification of particular features.

 While the hybrid strategy proposed is common practice in pertinent models (\cite{gha01int}), in the current framework it is also necessitated by the difficulty in sampling in the  high-dimensional state space of  $\bs{\Theta}$ (note that the projection matrices $\bs{P}^{(m)}$ in particular are of the dimension of the observables $d$ and $d>>1$) as well as the need for  scalability in the context of high-dimensional systems. The static parameters $\bs{\Theta}$ are estimated by maximizing the {\em log-posterior}.
\be
\label{eq:logpost}
L(\bs{\Theta})=\log \pi(\bs{\Theta} \mid \bs{y}_{1:\tau}) = \log \int \underbrace{\pi(\bs{\Theta}, \bs{\Theta}_{1:\tau} \mid \bs{y}_{1:\tau})}_{posterior ~\refeq{eq:post} }  ~d\bs{\Theta}_{1:\tau} 
\ee

Maximization of $L(\bs{\Theta})$ is  more complex than a standard optimization task as it involves integration over the unobserved dynamic variables $\bs{\Theta}_{1:\tau}$. While maximization can be accelerated by using gradient-based techniques (e.g. gradient ascent), the dimensionality of $\bs{\Theta}$ makes such an approach impractical as it can be extremely difficult to scale the parameter increments. We propose therefore adopting an Expectation-Maximization framework (EM) which is an iterative, robust scheme that is guaranteed to increase the log-posterior at each iteration (\cite{dem77max,gha01int}). It is based on constructing a series of increasing lower bounds of the log-posterior using auxiliary distributions $ q(\bs{\Theta}_{1:\tau})$:
\be
\label{eq:em}
\begin{array}{ll}
L(\bs{\Theta})=\log \pi(\bs{\Theta} \mid \bs{y}_{1:\tau}) &= \log \int \pi(\bs{\Theta}, \bs{\Theta}_{1:\tau} \mid \bs{y}_{1:\tau}) ~d\bs{\Theta}_{1:\tau}  \\
& = \log \int q(\bs{\Theta}_{1:\tau}) \frac{ \pi(\bs{\Theta}, \bs{\Theta}_{1:\tau} \mid \bs{y}_{1:\tau}) }{q(\bs{\Theta}_{1:\tau}) }  ~d\bs{\Theta}_{1:\tau} \\
& \ge \int q(\bs{\Theta}_{1:\tau}) \log \frac{ \pi(\bs{\Theta}, \bs{\Theta}_{1:\tau} \mid \bs{y}_{1:\tau}) }{q(\bs{\Theta}_{1:\tau}) }  ~d\bs{\Theta}_{1:\tau} \quad \textrm{(Jensen's inequality)} \\
&=F(q, \bs{\Theta}) 
\end{array}
\ee
It is obvious that this inequality becomes an equality when in place of the auxiliary distribution $q(\bs{\Theta}_{1:\tau})$ the posterior $\pi(\bs{\Theta}_{1:\tau} \mid  \bs{\Theta}, \bs{y}_{1:\tau})$ is selected. Given an estimate $\bs{\Theta}^{(s)}$ at step  $s$, this suggests iterating between an Expectation step (E-step) whereby  we average with respect to $q^{(s)}(\bs{\Theta}_{1:\tau})=\pi(\bs{\Theta}_{1:\tau} \mid  \bs{\Theta}^{(s)}, \bs{y}_{1:\tau})$ to evaluate the lower bound:
\be
\label{eq:estep}
\begin{array}{lll}
\textrm{\bf E-step:} & F^{(s)}(q^{(s)}, \bs{\Theta}) & =\int q^{(s)}(\bs{\Theta}_{1:\tau}) \log  \pi(\bs{\Theta}, \bs{\Theta}_{1:\tau} \mid \bs{y}_{1:\tau})   ~d\bs{\Theta}_{1:\tau} \\ & & - \int q^{(s)}(\bs{\Theta}_{1:\tau}) \log  q^{(s)}(\bs{\Theta}_{1:\tau}) ~d\bs{\Theta}_{1:\tau}
\end{array}
\ee
and a Maximization step (M-step) with respect to  $F^{(s)}(q^{(s)}, \bs{\Theta})$ (and in particular the first part in \refeq{eq:estep} since the second does no not depend on $\bs{\Theta}$):
\be
\label{eq:mstep}
\begin{array}{lll}
 \textrm{\bf M-step:}  & \bs{\Theta}^{(s+1)} &=arg \max_{\bs{\Theta}} F^{(s)}(q^{(s)}, \bs{\Theta}) \\
& & = arg \max_{\bs{\Theta}} E_{q^{(s)}(\bs{\Theta}_{1:\tau})}\left[ \log  \pi(\bs{\Theta}, \bs{\Theta}_{1:\tau} \mid \bs{y}_{1:\tau}) \right]  \\
& & = arg \max_{\bs{\Theta}} Q(\bs{\Theta}^{(s)}, \bs{\Theta}) \\
\end{array}
\ee


As the optimal auxiliary distributions $q^{(s)}(\bs{\Theta}_{1:\tau}) =\pi(\bs{\Theta}_{1:\tau} \mid  \bs{\Theta}^{(s)}, \bs{y}_{1:\tau})$ are intractable, we propose employing a Sequential Monte Carlo (SMC or particle filter, \cite{dou01seq,del06seq}) scheme for estimating the expectations  in the M-Step, i.e. \refeq{eq:mstep}. SMC samplers provide a {\em parallelizable} framework for  non-linear, non-Gaussian filtering problems whereby the target distribution $q^{(s)}(\bs{\Theta}_{1:\tau})=\pi(\bs{\Theta}_{1:\tau} \mid  \bs{\Theta}^{(s)}, \bs{y}_{1:\tau})$  is represented with  a population of $N$ particles $\bs{\Theta}^{(s,i)}_{1:\tau}$ and weights $W^{(s,i)}$ such that the expectation in \refeq{eq:mstep} can be approximated by:
\be
E_{q^{(s)}(\bs{\Theta}_{1:\tau})}\left[ \log  \pi(\bs{\Theta}, \bs{\Theta}_{1:\tau} \mid \bs{y}_{1:\tau}) \right] \approx \sum_{i=1}^N W^{(s,i)}~\log \pi(\bs{\Theta}, \bs{\Theta}^{(s,i)}_{1:\tau})  \mid \bs{y}_{1:\tau})
\ee
In section \ref{sec:rb} we discuss a particle filter that takes advantage of the particular structure  of the posterior and employs the locally optimal importance sampling distribution. Nevertheless,  SMC samplers involve  sequential importance sampling, and their performance decays with increasing  $\tau$ as the dimension of the state space $\bs{\Theta}_{1:\tau}$ increases  even when resampling and rejuvenation mechanisms are employed (\cite{and04par}).  Recent efforts based on exponential forgetting  have shown that the accuracy of the approximation can be improved  (while keeping the number of particles $N$ fixed) by employing {\em smoothing} (\cite{Godsill:2004}) over a fixed-lag in the past (\cite{cap05inf}).

In this paper we make use of an {\em approximate} but  highly efficient alternative proposed in \cite{and03onl,and04par,and05onl}. This is based on the so-called split-data likelihood (SDL) first discussed in \cite{ryd94con}, which consists of splitting the observations into blocks (overlapping or non-overlapping) of length $L < \tau$ and using the pseudo-likelihood which arises by assuming that these blocks are independent. It is shown in \cite{and04par} that this leads to an alternative Kullback-Leibler divergence contrast function and under some regularity conditions that the set of  parameters  optimizing this contrast function includes the true parameter. Because the size of the blocks is fixed, the degeneracy of particle filters can be averted and the quality of the approximations can be further improved by applying a backward smoothing step over each block (\cite{Godsill:2004}). Let $k$ denote the index of the block of length $L$  considered and  $\bar{\bs{y}}_k=\bs{y}_{(k-1)L+1:kL}$ and $\bar{\bs{\Theta}}_k=\bs{\Theta}_{(k-1)L+1:kL}$. If $\tau=r ~L$. The likelihood is  approximated by:
\be
p(\bs{y}_{1:\tau} \mid \bs{\Theta}, \bs{\Theta}_{1:\tau}) \approx \prod_{k=1}^r p(\bar{\bs{y}}_{k} \mid \bs{\Theta}, \bar{\bs{\Theta}}_{k})
\ee
When  $\bs{\Theta}_t$ has reached a stationary regime with invariant density $\nu_0(. \mid \bs{\Theta})$, then for any $k$,  $(\bs{\Theta}, \bar{\bs{\Theta}}_k,\bar{\bs{y}}_k)$ are identically distributed according to:
\be
\label{eq:sdlpost}
\begin{array}{ll}
\bar{p}(\bs{\Theta},\bar{\bs{\Theta}}_k,\bar{\bs{y}}_k)= & \pi(\bs{\Theta}) \nu_{0}(\bs{\Theta}_{(k-1)L} \mid \bs{\Theta}) p(\bs{y}_{(k-1)L} \mid  \bs{\Theta}_{(k-1)L}, \bs{\Theta}) \\
 &  \prod_{t=(k-1)L+1}^{kL-1} p(\bs{\Theta}_{t} \mid \bs{\Theta}_{t-1},\bs{\Theta}) p(\bs{y}_t \mid \bs{\Theta}_{t},  \bs{\Theta})
\end{array}
\ee

In a batch EM algorithm using the split-data likelihood and the $k^{th}$ block of data, the M-step would involve maximization with respect to $\bs{\Theta}$ of (see also \refeq{eq:mstep}):
\be
\label{eq:batch} 
\bar{Q}(\bs{\Theta}^{(k-1)}, \bs{\Theta})=\int \bar{p}(\bar{\bs{\Theta}}_k \mid \bs{\Theta}^{(k-1)}, \bar{\bs{y}}_k) \log \bar{p}(\bs{\Theta},\bar{\bs{\Theta}}_k,\bar{\bs{y}}_k) ~d\bs{\bar{\Theta}}_k
\ee
We utilize an {\em online} EM algorithm where the iteration numbers $s$ coincide with the block index $k$ (i.e. $s \equiv k$ ) which effectively implies that the estimates for $\bs{\Theta}$ are updated every time a new data block is considered.   The expectation step (E-step)  is replaced by a stochastic approximation (\cite{sat00onl,DBLP:journals/corr/abs-0712-4273}) while the M-step is left unchanged. In particular, at iteration  $k$ ($\equiv s$):
\be
\label{eq:estepsa} 
\begin{array}{lll}
 \textrm{ \bf online E-step} & \bar{Q}(\bs{\Theta}^{(1:k-1)}, \bs{\Theta}) & =(1-\gamma_k) \bar{Q}(\bs{\Theta}^{(1:k-2)}, \bs{\Theta}) \\
& &  +\gamma_k \int \bar{p} (\bar{\bs{\Theta}}_k \mid \bs{\Theta}^{(k-1)},  \bar{\bs{y}}_k) \log \bar{p}(\bar{\bs{\Theta}}_k,\bar{\bs{y}}_k) ~d\bar{\bs{\Theta}}_k 
\end{array}
\ee
and update the value of the parameters $\bs{\Theta}$ as:
\be
\begin{array}{ll}
\label{eq:mstepsa} 
 \textrm{\bf online M-step} &
\bs{\Theta}^{(k)} =arg \max_{\bs{\Theta}} \bar{Q}(\bs{\Theta}^{(1:k-1)}, \bs{\Theta})
\end{array}
\ee
The algorithm relies on a non-increasing sequence of positive stepsizes $\{ \gamma_k\}_{k\ge 0}$ such that $\sum_k \gamma_k = +\infty$ and $\sum_k \gamma_k^2 < +\infty$. In this work we adopted $\gamma_k=\frac{1}{k^a}$ with $a=0.51$.  Naturally the integrals above over the hidden dynamic variables $\bar{\bs{\Theta}}_k$ are estimated using SMC-based, particulate approximations of $\bar{p} (\bar{\bs{\Theta}}_k \mid \bs{\Theta}^{(k-1)} \bar{\bs{y}}_k)$. 
 For small $L$ the convergence will in general be slow as the split-block likelihood assumption will be further from the truth. For larger $L$, convergence is faster but the performance of the filter decays. For that purpose we also employed a backward smoothing filter  over each block using the algorithm described in \cite{Godsill:2004}. The computational cost of the smoothing algorithm is $O(N^2 L)$.

In practice, and in particular for the exponential distributions utilized in the proposed model (e.g. \refeq{eq:likec}), the EM iterations reduce to calculating  a set of (multivariate) sufficient statistics $\bs{\Phi}$. In particular, instead of the log-posterior lower bound $\bar{Q}(\bs{\Theta}^{(1:k-1)}, \bs{\Theta})$ in  \refeq{eq:estepsa} we update the sufficient statistics as follows:
\be
\label{eq:emss}
\begin{array}{ll}
 \bs{\Phi}^{(k)}& =(1-\gamma_k)\bs{\Phi}^{(k-1)} \\
& + \gamma_k \int \bar{p} (\bar{\bs{\Theta}}_k \mid \bs{\Theta}^{(k-1)}, \bar{\bs{y}}_k) \bs{\phi}(\bar{\bs{\Theta}}_k)~d\bar{\bs{\Theta}}_k
\end{array}
\ee
where $\int \bar{p} (\bar{\bs{\Theta}}_k \mid \bs{\Theta}^{(k-1)}, \bar{\bs{y}}_k) \bs{\phi}(\bar{\bs{\Theta}}_k)~d\bar{\bs{\Theta}}_k$ denotes the set of sufficient statistics associated with the block of data $\bs{\bar{y}}=\bs{{y}}_{(k-1)L+1:kL}$.
Specific details are provided in the Appendix. It is finally noted, that learning tasks in the context of the probabilistic model proposed,  should also involve identifying the correct {\em structure} (e.g. the number of different experts $M$). While this  problem poses some very challenging issues which are currently the topic of active research in various contexts (e.g. nonparametric methods), this paper is exclusively concerned with parameter learning. In section \ref{sec:examples}, we discuss Bayesian validation techniques for assessing quantitatively the correct model structure which are computationally feasible due to the efficiency  of the proposed algorithms.


\subsubsection{Locally optimal  Sequential Monte Carlo samplers}
\label{sec:rb}
In this section we discuss Monte Carlo approximations of the expectations appearing in \refeq{eq:emss} with respect to the density $\bar{p} (\bs{\bar{\Theta}}_{k} \mid \bs{\Theta}, \bs{y}_{1:L})={p} (\bs{\Theta}_{(k-1)L+1:kL} \mid \bs{\Theta}, \bs{y}_{(k-1)L+1:kL})$. Note that in order to simplify the notation we consider an arbitrary  block of length $L$ (e.g. $k=1$)and do not explicitly indicate the iteration number of the EM algorithm.
Hence the target density is:
\be
p (\bs{\Theta}_{1:L} \mid \bs{\Theta}, \bs{y}_{1:L})=\frac{1}{p(\bs{y}_{1:L} \mid \bs{\Theta} ) } \nu_0(\bs{\Theta}_1 \mid \bs{\Theta}) p(\bs{y}_1 \mid \bs{\Theta}_1,\bs{\Theta} ) \prod_{t=2}^L p(\bs{\Theta}_t \mid \bs{\Theta}_{t-1}, \bs{\Theta}) p(\bs{y}_t \mid \bs{\Theta}_t, \bs{\Theta})
\ee
where the dynamic variables are $\bs{\Theta}_t=( \bs{X}_t, \bs{z}_t)$ (\refeq{eq:xtot}).
Based on earlier discussions, the evolution dynamics of these variables are independent i.e.:
\be
\nu_0(\bs{\Theta}_1 \mid \bs{\Theta})=\nu_0(\bs{X}_1 \mid \bs{\Theta})v_0(\bs{z}_1 \mid \bs{\Theta}) 
\ee
and:
\be
 p(\bs{\Theta}_t \mid \bs{\Theta}_{t-1}, \bs{\Theta})= p(\bs{X}_t \mid \bs{X}_{t-1}, \bs{\Theta}) p(\bs{z}_t \mid \bs{z}_{t-1}, \bs{\Theta})
\ee
Since there is a deterministic relation between $\bar{\bs{z}}_t$ and $\bs{z}_t$ (\refeq{eq:ln}) we  use them interchangeably. In particular we use $\bs{\hat{z}}_t$ in the evolution equations since the initial and transition densities are  Gaussian (\refeq{eq:ouz}) and $\bs{z}_t$ in the likelihood densities as the expressions simplify  in \refeq{eq:obs}). The initial and transition densities for $\bs{X}_t$ are also Gaussian. Given that $\bs{x}_t^{(m)}$ are a priori independent, we have that:
\be
\begin{array}{ll}
p(\bs{X}_t \mid \bs{X}_{t-1}, \bs{\Theta})& =\prod_{m=1}^M p(\bs{x}_t^{(m)} \mid \bs{x}_{t-1}^{(m)}, \bs{\Theta}) \\
& = \mathcal{N}(\bs{X}_t \mid \bs{\mu}_t, \bs{S}_{X}) 
\end{array}
\ee
where the mean $\bs{\mu}_t=\bs{\mu}_t(\bs{X}_{t-1})$ is given by \refeq{eq:outrans} and $\bs{S}_{X}=\diag(\bs{S}_{x,1},\ldots,\bs{S}_{x,M})$ (from \refeq{eq:outrans} as well).

SMC samplers  operate on a sequence of target densities $p(\bs{\Theta}_{1:t} \mid \bs{y}_{1:t}, \bs{\Theta})$ which, for any $t$, are approximated by a  set of $n$ random samples (or {\em particles})  $\{ \bs{\Theta}_{1:t}^{(i)} \}_{i=1}^n$. These   are  propagated forward in time using a combination of {\em importance sampling},  {\em resampling}  and MCMC-based {\em rejuvenation} mechanisms \cite{mor06seq,del06seq,wea09var,vas08par}. Each of these particles is associated with an {\em importance  weight} $W^{(i)}$ ($\sum_{i=1}^n W^{(i)}=1$) which is  updated sequentially along with the particle locations in order to provide a particulate approximation:
\vspace{-.4cm}
\be                                                                             \label{eq:smc1}
p(\bs{\Theta}_{1:t} \mid \bs{y}_{1:t}, \bs{\Theta}) \approx \sum_{i=1}^n ~W^{(i)}~ \delta_{\bs{\Theta}_{1:t}^{(i)}} (\bs{\Theta}_{1:t}) 
\ee
where $\delta_{\bs{\Theta}^{(i)}_{1:t}}(.)$ is the Dirac function centered at $\bs{\Theta}^{(i)}_{1:t}$. Furthermore, for any  measurable   $\phi(\bs{\Theta}_{1:t}  )$  (as in \refeq{eq:emss}) and $\forall t$ \cite{mor04fey,Chopin:2004,cap05inf}:
\be 
\label{eq:smc2}
 \sum_{i=1}^n ~W^{(i)}~g(\bs{\Theta},\bs{\Theta}_{1:t} )  \rightarrow \int \phi(\bs{\Theta}_{1:t} )~p(\bs{\Theta}_{1:t} \mid \bs{y}_{1:t}, \bs{\Theta}) d\bs{\Theta}_{1:t} \quad \textrm{\em (almost surely)}
\ee

The particles are constructed recursively  in time using a sequence of importance sampling densities $q_t(\bs{\Theta}_t \mid \bs{\Theta}_{t-1}, \bs{y}_t, \bs{\Theta})$. The importance weights are determined from the fact that:
\be
p(\bs{\Theta}_{1:t} \mid \bs{y}_{1:t} , \bs{\Theta})=p(\bs{\Theta}_{1:t-1} \mid \bs{y}_{1:t-1} , \bs{\Theta}) \frac{ p(\bs{\Theta}_{t} \mid \bs{\Theta}_{t-1}, \bs{\Theta}) p(\bs{y}_t \mid \bs{\Theta}_{t} , \bs{\Theta}) }{ p(\bs{y}_t \mid \bs{y}_{1:t-1} , \bs{\Theta} ) }
\ee
Let $\{W^{(i)}, \bs{\Theta}^{(i)}_{1:t-1} \}_{i=1}^N$ the particulate approximation of  $p(\bs{\Theta}_{1:t-1} \mid \bs{y}_{1:t-1} , \bs{\Theta})$. Note that for $t=1$ and  for the Gaussian initial densities $\nu_0$ of the proposed model,  this consists of exact draws and weights $W^{(i)}=\frac{1}{N}$. At time $t$ we proceed as follows (\cite{dou01seq}):
\begin{enumerate}
\item Sample $\bs{\Theta}_t^{(i)} \sim q_t(\bs{\Theta}_t \mid \bs{\Theta}_{t-1}^{(i)}, \bs{y}_t, \bs{\Theta}), \forall i$ and set $\bs{\Theta}_{1:t}^{(i)} \leftarrow (\bs{\Theta}_{1:t-1}^{(i)},\bs{\Theta}_t^{(i)})$
\item Compute incremental weights:
\be
u_t^{(i)}=\frac{p(\bs{\Theta}_{t}^{(i)} \mid \bs{\Theta}_{t-1}^{(i)}, \bs{\Theta}) p(\bs{y}_t \mid \bs{\Theta}_{t}^{(i)} , \bs{\Theta}) }{q_t(\bs{\Theta}_t^{(i)} \mid \bs{\Theta}_{t-1}^{(i)}, \bs{y}_t, \bs{\Theta}) }
\ee
and update the weights:
\be
W^{(i)} = \frac{ W^{(i)} u_t^{(i)} }{ \sum_{j=1}^N W^{(i)} u_t^{(i)} }
\ee
\item Compute $ESS=\frac{1}{ \sum_{i=1}^N (W^{(i)})^2}$ and if $ESS< ESS_{min}$ perform multinomial resampling to obtain a new population with equally weighted particles ($ESS_{min}=N/2$ was used in this study). Set $t \leftarrow t+1$  and go to step 1.
\end{enumerate}

It can be easily established (\cite{dou00seq})  that the locally optimal importance sampling density is:
\be
q_t^{opt}(\bs{\Theta}_t \mid \bs{\Theta}_{t-1}, \bs{y}_t, \bs{\Theta})=\frac{p(\bs{\Theta}_{t} \mid \bs{\Theta}_{t-1}, \bs{\Theta}) p(\bs{y}_t \mid \bs{\Theta}_{t} , \bs{\Theta}) }{\int p(\bs{\Theta}_{t} \mid \bs{\Theta}_{t-1}, \bs{\Theta}) p(\bs{y}_t \mid \bs{\Theta}_{t} , \bs{\Theta}) ~d\bs{\Theta}_t } 
\ee
 In practice, it is usually impossible to sample from $q_t^{opt}$ and/or  calculate the integral in the denominator. As a result, approximations are used which nevertheless result in non-zero  variance in the  estimators. 
In this paper we take advantage of the fact that the transition density of $\bs{X}_t$ as well as the likelihood, conditioned on $\bs{z}_t$ are Gaussians and propose an importance sampling density of the form:
\be
q_t(\bs{X}_t, \bs{\hat{z}}_t \mid \bs{X}_{t-1},  \bs{\hat{z}}_{t-1}, \bs{y}_t, \bs{\Theta})=p(\bs{\hat{z}}_t \mid \bs{\hat{z}}_{t-1}, \bs{\Theta}) \frac{p(\bs{X}_{t} \mid \bs{X}_{t-1}, \bs{\Theta}) p(\bs{y}_t \mid \bs{X}_{t} , \bs{z}_t,  \bs{\Theta}) }{\int p(\bs{X}_{t} \mid \bs{X}_{t-1}, \bs{\Theta}) p(\bs{y}_t \mid \bs{X}_{t} , \bs{z}_t,  \bs{\Theta}) ~d\bs{X}_t }
\ee
This implies using the prior to draw $\bs{\hat{z}}_t$ and the locally optimal distribution (conditioned on $\bs{\hat{z}}_t$ or equivalently $\bs{z}_t$) for $\bs{X}_t$. The latter will also be a Gaussian whose mean $\bar{\bs{\mu}}_t$ and covariance $\bs{\bar{S}}_X$ can be readily be established (e.g. using Kalman filter formulas):
\be
\begin{array}{l}
\bs{\bar{S}}_X=\left( \bs{S}_{X}^{-1}+\bs{W}_t^T\bs{\Sigma}^{-1} \bs{W}_t \right)^{-1} \\
\bar{\bs{\mu}}_t=\bs{\bar{S}}_X \left( \bs{S}_{X}^{-1}\bs{\mu}_t+ \bs{W}_t^T\bs{\Sigma}^{-1} \bs{y}_t \right)
\end{array}
\ee

As a result the incremental weights $u_t$ are given by:
\be
u_t=\mid \bs{\bar{S}}_X \mid ^{1/2} \exp\{ \frac{1}{2} \bar{\bs{\mu}}_t^T \bs{\bar{S}}_X^{-1}\bar{\bs{\mu}}_t-\frac{1}{2} \bs{\mu}_t^T \bs{S}_{X}^{-1}\bs{\mu}_t \}
\ee




\subsection{Prediction and Bayesian adaptive time-integration}
\label{sec:bayesint}
%
 Bayesian inference results do not include just point estimates  but rather samples from the posterior density, at least with repsect to the time-varying parameters $\bs{\Theta}_t$. 
The inferred posterior can be readily used to make {\em probabilistic predictions} about the future evolution of the high-dimensional, multiscale process $\bs{y}_t$. Given observations $\bs{y}_{1:\tau}=\{ \bs{y}_t \}_{t=1}^{\tau}$, the {\em predictive posterior} for the future state  of the system $\bs{y}_{\tau+1:\tau+T}$ over a time horizon $T$ can be expressed as:
\begin{eqnarray}
 \label{eq:pred}
p(\bs{y}_{\tau+1:\tau+T} \mid \bs{y}_{1:\tau}) & = & \int p(\bs{y}_{\tau+1:\tau+T}, \bs{\Theta}, \bs{\Theta}_{\tau+1:\tau+T} \mid \bs{y}_{1:\tau})~ d\bs{\Theta} d\bs{\Theta}_{\tau+1:\tau+T} \nonumber \\
& = & \int \underbrace{ p(\bs{y}_{\tau+1:\tau+T} \mid \bs{\Theta}, \bs{\Theta}_{\tau+1:\tau+T} ) }_{likelihood ~\refeq{eq:like}} ~p(\bs{\Theta}, \bs{\Theta}_{\tau+1:\tau+T} \mid \bs{y}_{1:\tau})  d\bs{\Theta}~d\bs{\Theta}_{\tau+1:\tau+T} \nonumber \\
& = & \int p(\bs{y}_{\tau+1:\tau+T} \mid \bs{\Theta}, \bs{\Theta}_{\tau+1:\tau+T} ) \nonumber \\
& & ~\underbrace{ p(\bs{\Theta}_{\tau+1:\tau+T} \mid \bs{\Theta}_{\tau}, \bs{\Theta})}_{prior ~\refeq{eq:priortot}}~   \underbrace{ p(\bs{\Theta}, \bs{\Theta}_{\tau} \mid \bs{y}_{1:\tau}) }_{posterior ~\refeq{eq:post}} d\bs{\Theta}~d\bs{\Theta}_{\tau:\tau+T}
\end{eqnarray}

The integral above can be approximated using Monte Carlo. In particular given the particulate approximation of the posterior $p(\bs{\Theta}, \bs{\Theta}_{1:\tau} \mid \bs{y}_{1:\tau})$ (which consists of samples of the dynamic variables $\bs{\Theta}_t$ and the MAP estimate of $\bs{\Theta}$), samples from the prior  $p(\bs{\Theta}_{\tau+1:\tau+T} \mid \bs{\Theta}_{\tau}, \bs{\Theta})$ and subsequently the likelihood $p(\bs{\Theta}_{\tau+1:\tau+T} \mid \bs{\Theta}_{\tau}, \bs{\Theta})$ can readily be drawn. In fact, given that the latter is  a multivariate Gaussian, the predictive posterior will consist of a mixture of Gaussians, one for each sample of  $\bs{\Theta}_{\tau+1:\tau+T}$ drawn.

\begin{figure}[!t]
\centering
\psfrag{prior}{prior}
\psfrag{p1}{$p(\bs{\Theta}, \bs{\Theta}_{1:\tau})$}
\psfrag{like}{likelihood}
\psfrag{p2}{$p( \bs{y}_{1:\tau} \mid \bs{\Theta}, \bs{\Theta}_{1:\tau})$ }
\psfrag{comp}{computational}
\psfrag{exp}{\em experimental}
\psfrag{data}{data $\bs{y}_{1:\tau}$}
\psfrag{datae}{\em data $\bs{y}_{1:\tau'}$}
\psfrag{fine}{\bf fine-scale}
\psfrag{coarse}{\bf coarse-grained}
\psfrag{post}{posterior}
\psfrag{p3}{$p(\bs{\Theta}, \bs{\Theta}_{1:\tau} \mid \bs{y}_{1:\tau})$}
\psfrag{pred}{predictive post.}
\psfrag{p4}{$p(\bs{y}_{\tau+1:\tau+T} \mid \bs{y}_{1:\tau})$}
\psfrag{cons}{consistent}
\psfrag{reinit}{reinitialization}
\psfrag{model}{\bf model}
\psfrag{postmean}{posterior mean}
\psfrag{cred99}{$99\%$ post. quantile}
\psfrag{cred1}{$1\%$ post. quantile}
\psfrag{acce}{\em acceptable predictive}
\psfrag{acee2}{\em uncertainty}
\psfrag{y1}{$y^{(1)}$}
\psfrag{y2}{$y^{(2)}$}
\psfrag{bayes}{\bf \em Bayesian}
\includegraphics[width=0.86\textwidth,height=8cm]{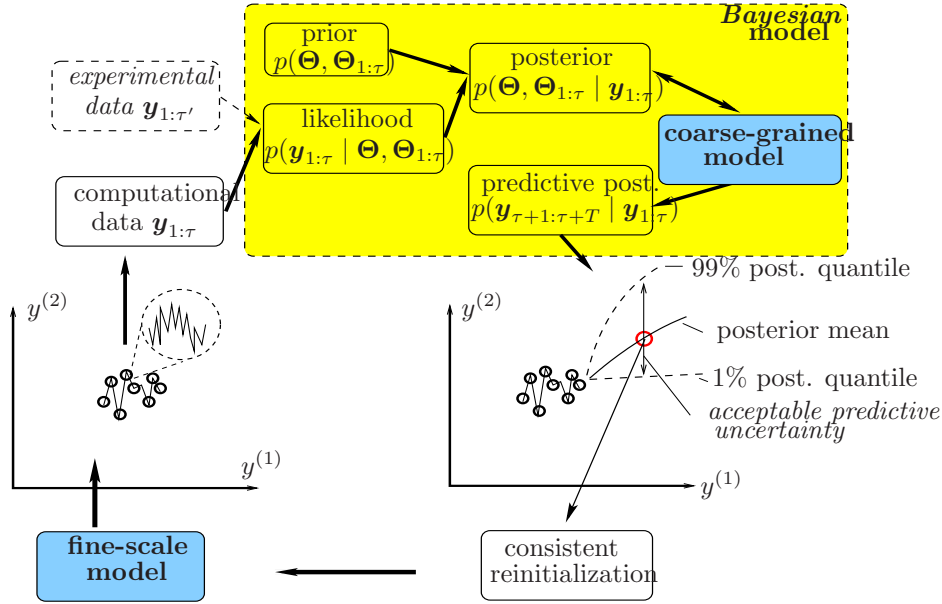}
\caption{\em 
Bayesian adaptive time-integration and data-model fusion illustrated for a two-dimensional flow. The data generated from computational simulations $\bs{y}_{1:\tau}$ and/or experiments $\bs{y}_{1:\tau'}$ are {\em sequentially} incorporated in the Bayesian model and the posterior $p(\bs{\Theta}, \bs{\Theta}_{1:\tau} \mid \bs{y}_{1:\tau})$ over dynamic and static parameters is updated. The predictive posterior $p(\bs{y}_{\tau+1:\tau+T} \mid \bs{y}_{1:\tau})$ is over the time horizon $T$ used to efficiently produce probabilistic predictions of the evolution  of the high-dimensional process $\bs{y}_t$ in the future. When the uncertainty associated with those predictions exceeds the analysts' tolerance, the original system is {\em consistently} reinitialized and more data are generated. These are used to update the (predictive) posterior and to produce additional predictive estimates. It is noted that the tolerance in the predictive uncertainty can also be measured with respect to (low-dimensional)  observables which are usually of interest in practical applications.
}
\label{fig:pan}
\end{figure}

The important consequence of the Bayesian framework advocated is that precitive estimates are not restricted to point estimates but whole distributions which can readily quantify the predicitve uncertainty. 
This naturally gives rise  to  Bayesian, adaptive, time-integration scheme that allows bridging across timescales while providing quantitative, probabilistic estimates of the accuracy of the coarse-grained dynamics (Figure \ref{fig:pan}). The distribution of \refeq{eq:pred} is used to probabilistically predict the evolution of the system. The time range over which the reduced model is employed does not have to be specified a priori but can be automatically determined by the variance of the predictive posterior (Figure  \ref{fig:pan}). Once this exceeds the allowable tolerance specified by the analyst, the fine-scale process is reinitialized and more data are obtained, that can in turn be used to update the coarse-grained model. It is emphasized  that due to the generative character of the model proposed, the reinitialization can be performed consistently based in general on the emission equations  \refeq{eq:obs}. In contrast to existing techniques such as projective and coarse-projective integration \cite{Gear:2003a,Gear:2002,Gear:2003,Gear:2005,Kavousanakis:2007,RicoMartinez:2004} as well as Heterogeneous Multiscale Methods \cite{e05ana,li08eff}, there is no need to prescribe  {\em lifting} and  {\em restriction} operators and no ambiguity exists with regards to the appropriateness of the reinitialization scheme. Furthermore, the probabilistic coarse-grained model provides quantitative estimates for its predictive ability and automatically identifies the need for more information from the fine-scale model.

\section{Numerical experiments}
\label{sec:examples}
The first examples is intended to validate the accuracy of the proposed online EM scheme and utilizes a synthetic dataset. The second example uses actual data and illustrates the superiority of the proposed PMLDS model over existing SLDS models. Finally the third example provides an application in the context of multiscale simulations for the time-dependent diffusion equation.

\subsection{Synthetic data}
\begin{figure}[!t]
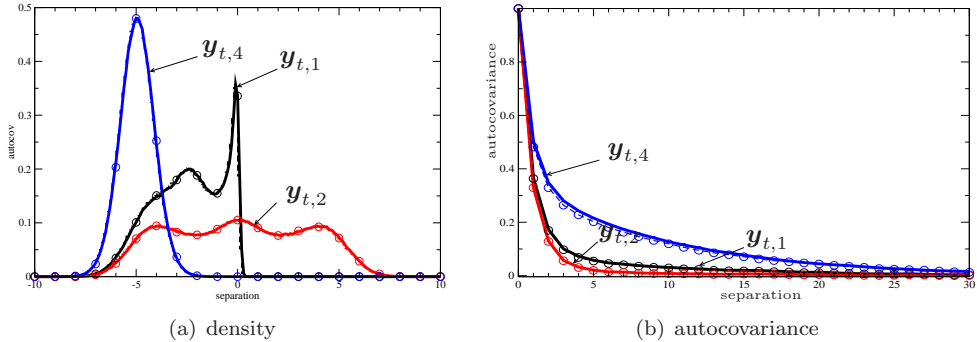

\vspace{.5cm}
\subfigure[density]{
\label{fig:exa_den}
\psfrag{y1}{$\bs{y}_{t,1}$}
\psfrag{y2}{$\bs{y}_{t,2}$}
\psfrag{y4}{$\bs{y}_{t,4}$}
\includegraphics[width=0.45\textwidth,height=4cm]{FIGURES/exa_compare_nsamp=200_den.eps}} \hfill
\subfigure[autocovariance]{
\label{fig:exa_cov}
\psfrag{y1}{$\bs{y}_{t,1}$}
\psfrag{y2}{$\bs{y}_{t,2}$}
\psfrag{y4}{$\bs{y}_{t,4}$}
\psfrag{autocov}{\tiny autocovariance}
\psfrag{separation}{\tiny separation}
\includegraphics[width=0.5\textwidth,height=4cm]{FIGURES/exa_compare_nsamp=200_cov.eps}}
\caption{\em Densities (a)) and Autocovariances (b)) of times series $\bs{y}_{t,1}$ (black),  $\bs{y}_{t,2}$ (red) and $\bs{y}_{t,4}$ (blue) (solid lines). With ($- \circ -)$ the densities and autocovariances of the same times series generated using the learned model parameters using the proposed online EM scheme with $L=200$ and $N=200$ (see Figures \ref{fig:exa_b} and  \ref{fig:exa_error}) }
\vspace{.5cm}
\label{fig:exa}
\end{figure}

We generated data from the proposed model in order to investigate the ability of the inference and learning algorithms discussed. In particular, we considered a mixture of two $M=2$, one-dimensional OU processes ($K=1$) as in \refeq{eq:oux} with $(b^{(1)},\bs{q}_x^{(1)},\bs{\Sigma}^{(1)})=(0.1,-5.0,0.2)$ (slow) and $(b^{(2)},\bs{q}_x^{(2)},\bs{\Sigma}^{(2)})=(1.,+5.0,2.0)$ (fast). The logistic normal distribution was used to model the weights associated with each of the hidden processes using  an  isotropic Ornstein-Uhlenbeck process $d\mathbf{z}_t=-b_z (\mathbf{z}_t-\bs{q}_z)dt+\bs{\Sigma}_z^{1/2} d\bs{W}_t$ with $b_z=1.0$, $\bs{q}_z=[0,~0]^T$ and $\bs{\Sigma}_z=\left[  \begin{array}{cc}  10. & 0  \\ 0 & 10. \end{array}   \right]$. Two $10\times 1$ projection vectors $\bs{P}^{m}, m=1,2$ were generated from the prior $\mathcal{N}(\bs{0}, 100 \bs{I})$ (see Appendix) and ($d=10$)  time series $\bs{y}_t$ were produced based on \refeq{eq:obs} with idiosyncratic variances $\bs{\Sigma}=0.1^2 \bs{I}$ and time step $\delta t=1$. 
The resulting time series exhibit multimodal, non-Gaussian densities as can be seen in Figure \ref{fig:exa_den} as well as two distinct time scales as it can be seen in the autocovariances plotted in Figure \ref{fig:exa_cov}.

\begin{figure}{!t}
\psfrag{b1}{$b_x^{(1)}$}
\psfrag{b2}{$b_x^{(2)}$}
\psfrag{observations}{time $t$}
 \includegraphics[width=0.75\textwidth,height=5cm]{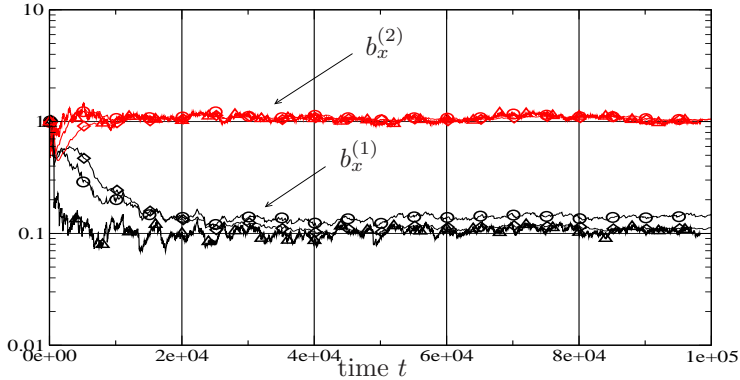}
\caption{Convergence of $b_x^{(1)}$ (black) and $b_x^{(2)}$ (red) using the online EM algorithm for three different combinations of $L$ and $N$. ($- \circ -$) corresponds to $L=100$, $N=100$, ($- \Diamond -$) to $L=200$, $N=200$ and ($- \triangle -$) to $L=20$, $N=1000$}
\label{fig:exa_b}
\end{figure}

\begin{figure}[!t]
\psfrag{refer}{ref. value $18.03$}
\psfrag{b2}{$b_x^{(2)}$}
\psfrag{observations}{time $t$}
 \includegraphics[width=0.75\textwidth,height=5cm]{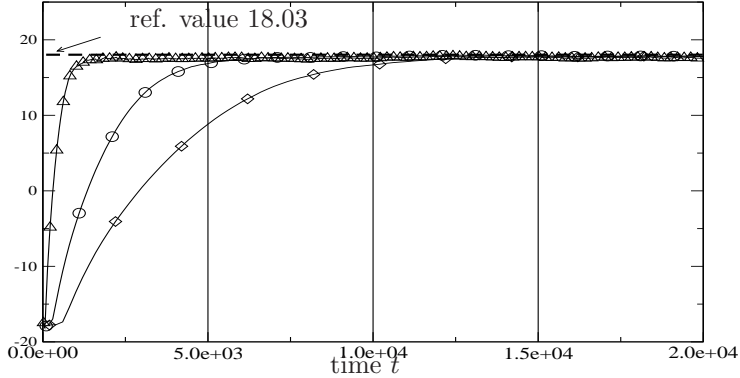}
\caption{Log-likelihood convergence using the online EM algorithm for three different combinations of $L$ and $N$. ($- \circ -$) corresponds to $L=100$, $N=100$, ($- \Diamond -$) to $L=200$, $N=200$ and ($- \triangle -$) to $L=20$, $N=1000$}
\label{fig:exa_ll}
\end{figure}

\begin{figure}[!t]
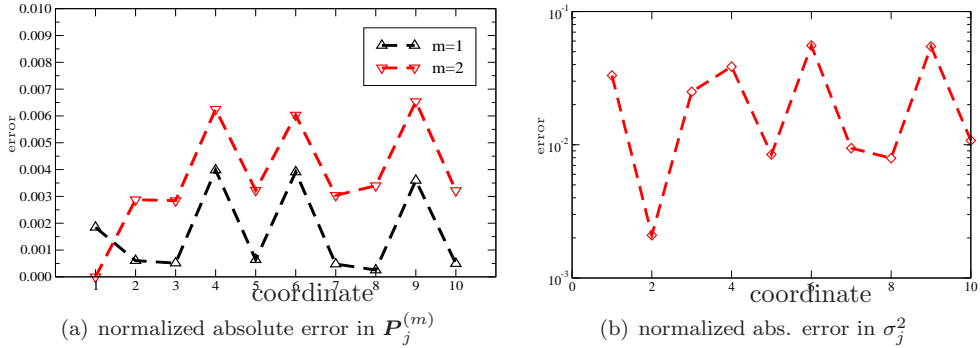

\vspace{.25cm}
\subfigure[normalized absolute error in $\bs{P}^{(m)}_j$]{
 \label{fig:exa_P}
\psfrag{coordinate}{coordinate}
\psfrag{error}{\tiny error}
\includegraphics[width=0.5\textwidth,height=4cm]{FIGURES/exa_compare_nsamp=200_error_Tji.eps}}\hfill
\subfigure[normalized abs. error in $\sigma_j^2$]{
\label{fig:exa_sj2}
\psfrag{coordinate}{coordinate}
\psfrag{error}{\tiny error}
\includegraphics[width=0.45\textwidth,height=4cm]{FIGURES/exa_compare_nsamp=200_error_sj2.eps}}
\caption{\em Normalized absolute errors (per coordinate) on the identification of the projection vectors $\bs{P}^{(m)}, m=1,2$ and idiosyncratic variances $\sigma_j^2, j=1,\ldots,d$ using the proposed online EM scheme with $L=200$ and $N=200$}
\label{fig:exa_error}
\end{figure}

Figure \ref{fig:exa_b} depicts the convergence of the proposed online EM scheme to the reference values of $b_x^{(m)}, m=1,2$, for various block sizes $L$ and particle populations $N$. Figure \ref{fig:exa_ll} depicts the evolution of the log-likelihood per iteration of the EM algorithm. Figure \ref{fig:exa_P} depicts the normalized error in the identified $\bs{P}^{(m)}, m=1,2$ and isiosyncratic variances  $\bs{\Sigma}$ pre coordinate after 20,000 iterations.   In all cases  the algorithm exhibits good convergence to the reference values.

\subsection{Temperature Dataset}

The goal of this numerical experiment is to illustrate the interpretability of the proposed model and compare with the switching-state linear model discussed in section \ref{sec:genintro} (\refeq{eq:obs_slds}). For that purpose we utilized the temperature data (in degrees Fahrenheit) of the capitals of the 50 states in the U.S.A ($d=50$). The data was obtained from {\em  http://www.engr.udayton.edu/weather/citylistUS.htm} and it represents the average daily temperatures from 01/01/1996 until  01/13/2009 (i.e. $5,127$ daily observations).

\begin{figure}[!t]
\psfrag{refer}{ref. value $18.03$}
\psfrag{b2}{$b_x^{(2)}$}
\psfrag{observations}{time $t$}
\includegraphics[width=0.75\textwidth,height=6cm]{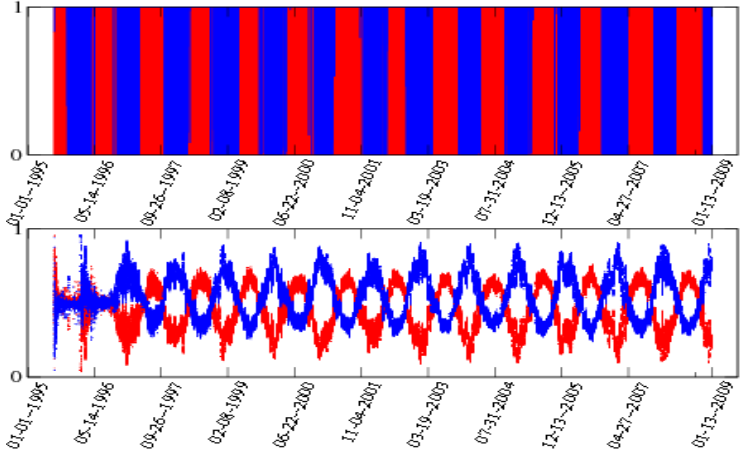}
\caption{(Top) Posterior mean of $z_{m,t}, m=1,2$ based on the SLDS and (Bottom) particulate approximation of the posterior of $z_{m,t}, m=1,2$  PMLDS. Both results were obtained using the previously discussed online EM scheme with $L=200$ and $N=100$}
\label{fig:exb_post_memb}
\end{figure}

\begin{figure}[!t]
\psfrag{hot}{hot/summer}
\psfrag{cold}{cold/winter}
\psfrag{observations}{time $t$}
\includegraphics[width=0.75\textwidth,height=5cm]{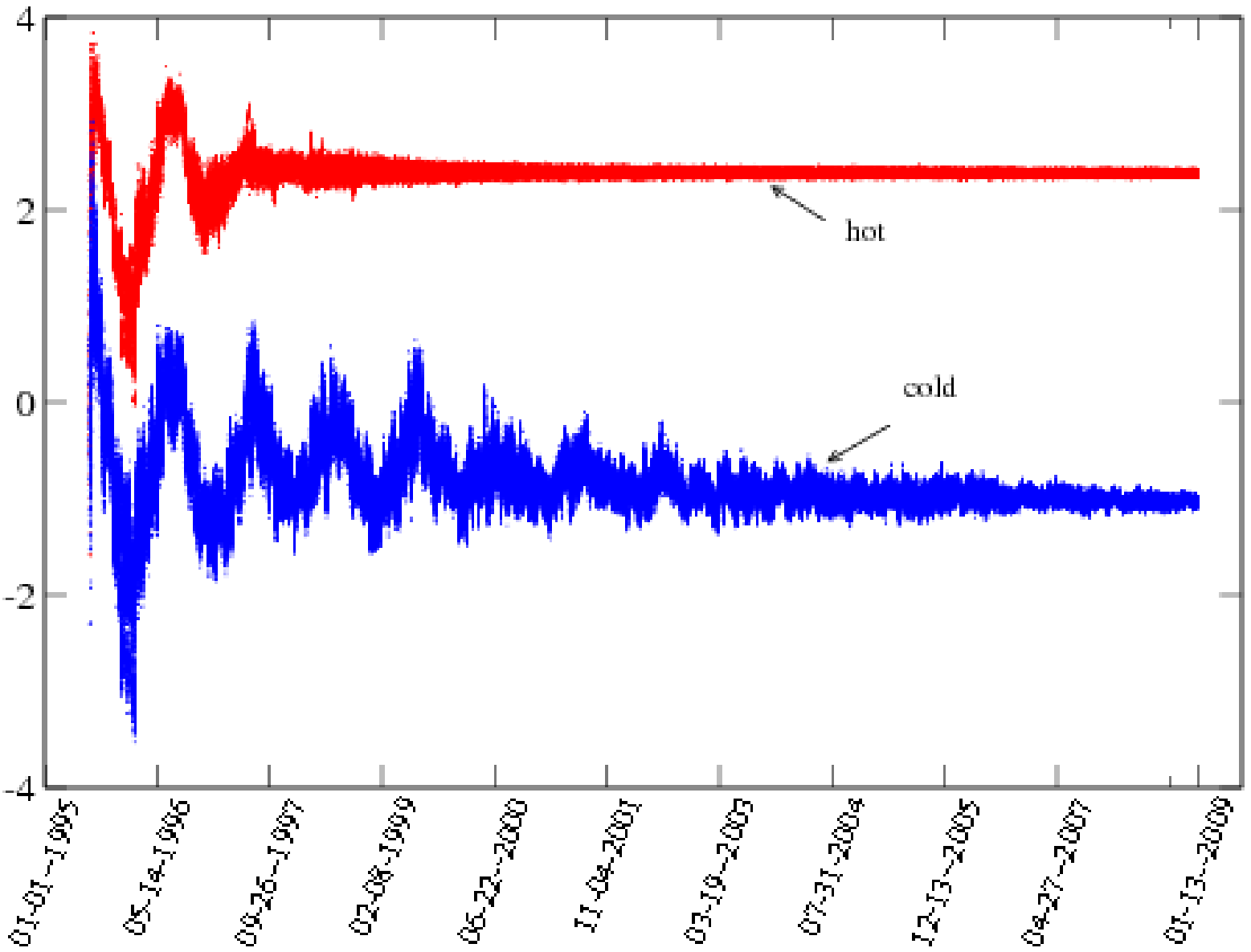}
\caption{Particulate approximation of the posterior of $\bs{x}_t^{(m)}, m=1,2$. Results were obtained using the previously discussed online EM scheme with $L=200$ and $N=100$}
\label{fig:exb_post_hidd}
\end{figure}

\begin{figure}[!ht]
\psfrag{refer}{ref. value $18.03$}
\psfrag{b2}{$b_x^{(2)}$}
\psfrag{observations}{time $t$}
\includegraphics[width=0.85\textwidth,height=8cm]{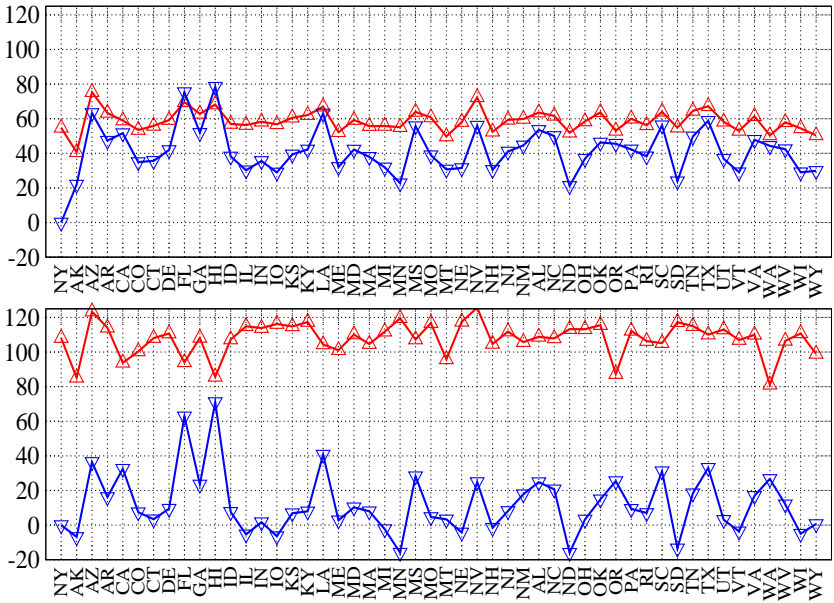}
\caption{Maximum posterior estimates of $\bs{P}^{(m)}\in \RR^{50}, m=1,2$ based on the SLDS(top) and PMLDS (bottom) models.  Both results were obtained using the previously discussed online EM scheme with $L=200$ and $N=100$}
\label{fig:exb_eigen}
\end{figure}

Figure \ref{fig:exb_post_memb} depicts the posterior memberships corresponding to the SLDS and PMLDS models based on a reduced model with two hidden states ($M=2$) described by one-dimensional OU processes ($K=1$). The former assumes that at each time instant  the observables $\bs{y}_t$ arises from a {\em single} hidden process. Hence  a single entry of  $\bs{z}_t=[z_{1,t}, z_{2,t}, \ldots, z{M,t}]$ is equal to 1 and the rest are all equal to 0. The top part of  Figure \ref{fig:exb_post_memb} shows the posterior mean of $z_{m,t},  m=1,2$.  As one would expect  the two-states correspond to cold-winter (blue)  and hot-summer (red) and alternate periodically (roughly the cold-winter state is active between early November until mid-April and   the hot-summer state in the remainder of the calendar year). The top part of Figure \ref{fig:exb_eigen} depicts the corresponding $\bs{P}^{(m)}, m=1,2$ where southern states have higher values and northern states lower.
Naturally, winter and summer represent the two extremes but several intermediate states are also present. The proposed partial-membership model can account for those  states without increasing the cardinality of the reduced-order dynamics. As it can be seen in the bottom part of Figure \ref{fig:exb_post_memb} which depicts the particulate approximation of the posterior of $z_{m,t},  m=1,2$, the two hidden states can also be attributed to the two extremes but the actual temperatures arise by a weighted combination of these two. Naturally during spring-summer months the weight of the ``red'' state is higher and during autumn-winter months the weight of the ``blue'' state takes over. The posterior of the hidden processes $\bs{x}_t^{(m)}, m=1,2$ is depicted in Figure \ref{fig:exb_post_hidd}.

In order to quantitatively compare the two models we calculated the average, one-step-ahead, predictive log-likelihood $\log p(\bs{y}_{t+1} \mid \bs{y}{1:t}), \forall t \in [0,T)$:
\begin{eqnarray}
 \label{eq:pred_exb}
p(\bs{y}_{t+1} \mid \bs{y}_{1:t}) & = & \int \log p(\bs{y}_{t+1} \mid  \bs{\Theta}, \bs{\Theta}_{t+1}  \bs{y}_{1:t})~p(\bs{\Theta}, \bs{\Theta}_{t+1} \mid \bs{y}_{1:t}) d\bs{\Theta} d\bs{\Theta}_{t+1} \nonumber \\
& = & \int \log p(\bs{y}_{t+1} \mid  \bs{\Theta}, \bs{\Theta}_{t+1}  \bs{y}_{1:t}) \nonumber \\
&  &~\underbrace{ p(\bs{\Theta}_{t+1} \mid \bs{\Theta}_{t}, \bs{\Theta})}_{prior}~   \underbrace{ p(\bs{\Theta}, \bs{\Theta}_{t} \mid \bs{y}_{1:t}) }_{posterior} d\bs{\Theta}~d\bs{\Theta}_{t:t+1}
\end{eqnarray}
The latter integral is approximated by Monte Carlo  using the MAP estimate of $\bs{\Theta}$ and the particulate approximation of the posterior for the dynamic variables $ \bs{\Theta}_{t}$.  This provides a measure of how well the model generalizes to a novel observation from the same distribution as the training data and higher values imply a better model. Table \ref{tab:exb} reports the average values (in bits) plus/minus the standard deviation (over $t \in (100,T=5127)$ ). Similar calculations were carried out for other model cardinalities (i.e. $M, K$) and in all cases the proposed model exhibited superior performance. This superiority becomes more pronounced as $M$ and $K$ increased which can be attributed to the factorial  character of PMLDS.

\begin{table}[!t]
 \begin{tabular}{|c|c|r|r|}
  \hline \hline
$M$ & $K$ & SLDS & PMLDS \\
\hline
$2$ & $1$ & $-179.97 \pm 37.31$ & $-171.11 \pm 37.20$  \\
$2$ & $2$ & $-170.68 \pm 36.95$ & $-141.11  \pm 27.82$  \\
$4$ & $1$ & $-176.40 \pm 34.36$ & $-143.81 \pm 25.56$  \\
$4$ & $2$ & $-166.05 \pm 30.57$ & $-117.67 \pm 21.15$  \\
\hline \hline 
 \end{tabular}
\caption{One-step-ahead predictive log-likelihood (\refeq{eq:pred_exb}) of SLDS and PMLDS models for various $M$, $K$. The table reports the average value plus/minus one standard deviation in bits. All results were obtained using the previously discussed online EM scheme with $L=200$ and $N=100$}
\label{tab:exb}
\end{table}

\subsection{Transient Heat equation}

\begin{figure}[!t]
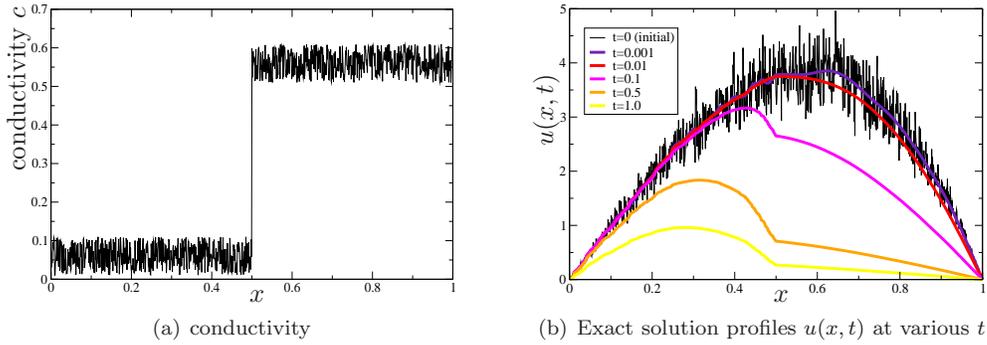

\subfigure[conductivity]{
 \label{fig:exc_cond}
\psfrag{x}{$x$}
\psfrag{conductivity}{conductivity $c$}
\includegraphics[width=0.45\textwidth,height=4cm]{FIGURES/material_properties.eps}}\hfill
\subfigure[Exact solution profiles $u(x,t)$ at various $t$]{
\label{fig:exc_solution}
\psfrag{x}{$x$}
\psfrag{temp}{ $u(x,t)$}
\includegraphics[width=0.45\textwidth,height=4cm]{FIGURES/temperature_profile.eps}}
\caption{\em Dynamic heat equation. Spatial discretization with $1,000$ finite elements. Time discretization using $\delta t=1\times 10^{-4}$. }
\label{fig:exc_config}
\end{figure}

  We finally apply the proposed analysis scheme to the one-dimensional transient heat equation:
\be
\label{eq:he}
\left\{ 
\begin{array}{l}
\frac{\pa u}{\pa t}=\frac{\pa }{\pa x } \left(a(x) \frac{\pa u}{\pa x} \right), \quad x \in [0,1] \\
u(0,t)=u(1,t)=0, \quad \forall t
\end{array}
\right.
\ee
The spatial domain was discretized with $1,000$ finite elements of equal length and we  considered a ``rough'' conductivity profile shown in Figure \ref{fig:exc_cond}. The conductivity $a(x)$ in each finite element was assumed constant and its value was drawn independently from a uniform distribution \footnote{We considered a single realization of the conductivity profile and solved for it as a deterministic problem. The stochastic PDE where $a(x)$ is random is not considered here.}.
For $x\in [0,0.5]$ we used the uniform $U[0.01,0.1]$ and for $x\in (0.5, 1]$, $U[0.51,0.6]$. This naturally resulted in the jump observed in Figure \ref{fig:exc_cond} which as a consequence gave rise to two distinct slow time scales in the solution profile $u(x,t)$ depicted in Figure \ref{fig:exc_solution}.
A rough profile of initial conditions was also used (as can be seen Figure \ref{fig:exc_solution}, $t=0$). In particular at each node $x_i=0.001 i,~i=1,\ldots,1001$ we set $u(x_i,0)=10x_i(1-x_i)(1+0.1 Z_i)$ where $Z_i \sim N(0,1)$ (i.i.d).

\begin{figure}[!t]
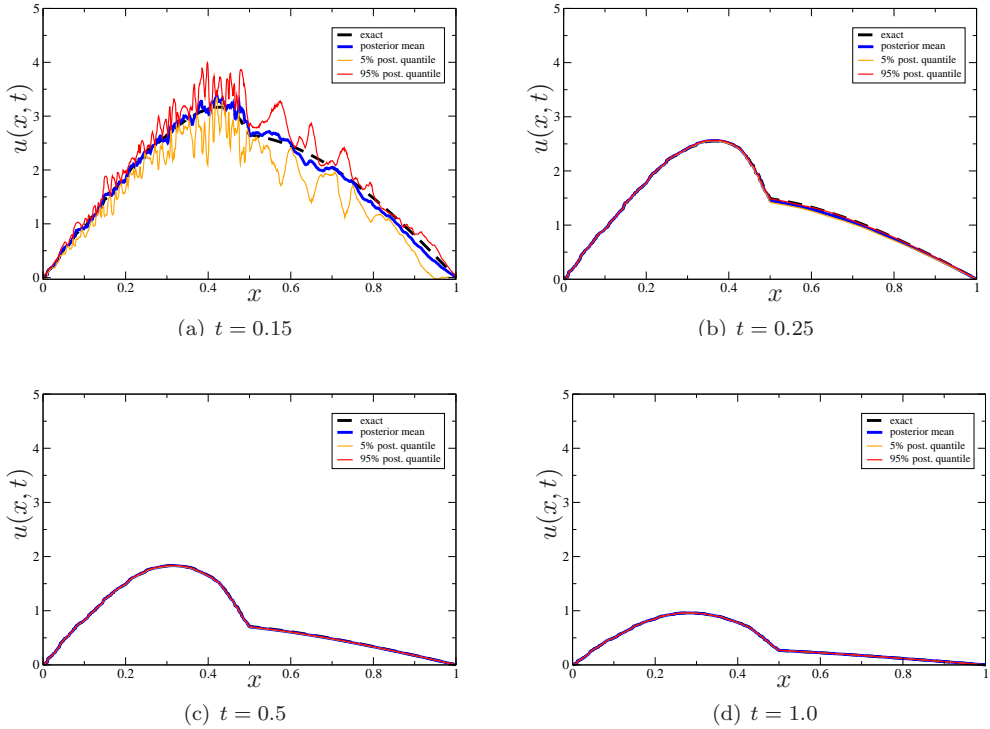

\subfigure[$t=0.15$]{
 \label{fig:exc_1500}
\psfrag{x}{$x$}
\psfrag{temp}{$u(x,t)$}
\includegraphics[width=0.45\textwidth,height=4cm]{FIGURES/posterior_slice_1500.eps}}
\hfill
\subfigure[$t=0.25$]{
\label{fig:exc_2500}
\psfrag{x}{$x$}
\psfrag{temp}{$u(x,t)$}
\includegraphics[width=0.45\textwidth,height=4cm]{FIGURES/posterior_slice_2500.eps}}
\vspace{0.25cm} \\
\subfigure[$t=0.5$]{
 \label{fig:exc_5000}
\psfrag{x}{$x$}
\psfrag{temp}{$u(x,t)$}
\includegraphics[width=0.45\textwidth,height=4cm]{FIGURES/posterior_slice_5000.eps}}
\hfill
\subfigure[$t=1.0$]{
\label{fig:exc_10000}
\psfrag{x}{$x$}
\psfrag{temp}{$u(x,t)$}
\includegraphics[width=0.45\textwidth,height=4cm]{FIGURES/posterior_slice_10000.eps}}
\caption{\em Comparison of predictive posterior estimates (posterior mean and $5\%$ and $95\%$ quantiles) with exact solution $u(x,t)$ at various $t$ }
\label{fig:exc_posterior}
\end{figure}

\begin{figure}[!t]
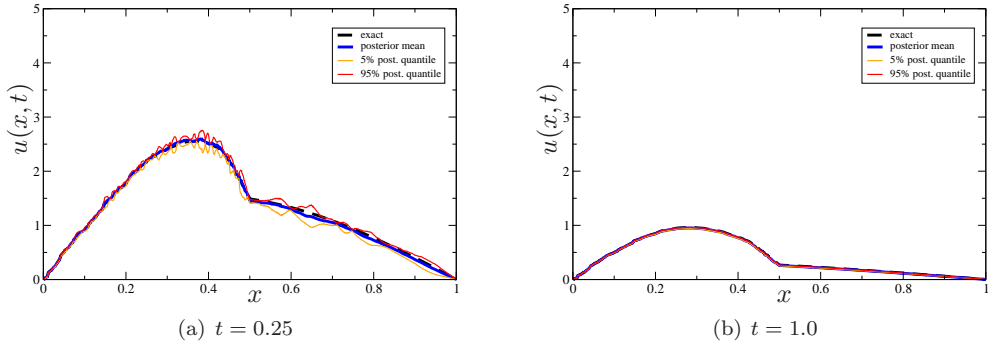

\subfigure[$t=0.25$]{
 \label{fig:exc_2500_2}
\psfrag{x}{$x$}
\psfrag{temp}{$u(x,t)$}
\includegraphics[width=0.45\textwidth,height=4cm]{FIGURES/posterior_slice_2500_ntpred=1000.eps}}
\hfill
\subfigure[$t=1.0$]{
\label{fig:exc_10000_2}
\psfrag{x}{$x$}
\psfrag{temp}{$u(x,t)$}
\includegraphics[width=0.45\textwidth,height=4cm]{FIGURES/posterior_slice_10000_ntpred=1000.eps}}
\caption{\em Comparison of predictive posterior estimates (posterior mean and $5\%$ and $95\%$ quantiles) with exact solution $u(x,t)$ at various $t$. These results were obtained with a prediction horizon $T=0.1$ ($\delta t =0.0001$) in contrast to Figure \ref{fig:exc_posterior} which were obtained for $T=0.05$.}
\label{fig:exc_posterior_2}
\end{figure}

\begin{figure}[!t]
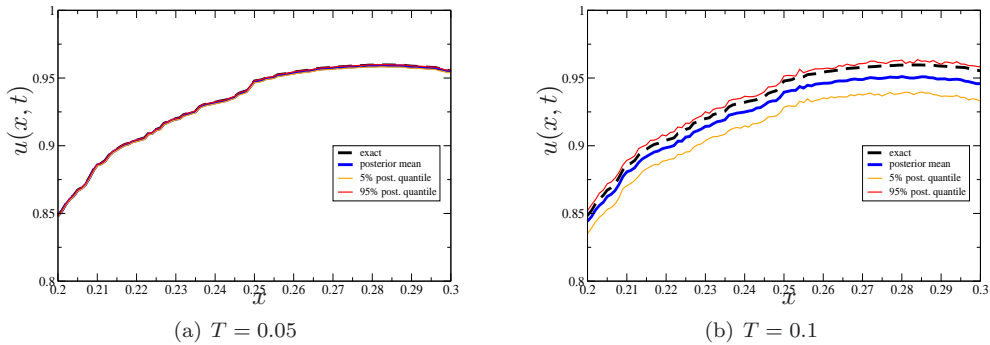

\subfigure[$T=0.05$]{
 \label{fig:exc_10000_3}
\psfrag{x}{$x$}
\psfrag{temp}{$u(x,t)$}
\includegraphics[width=0.45\textwidth,height=4cm]{FIGURES/posterior_slice_10000_zoom_x=0.25.eps}}
\hfill
\subfigure[$T=0.1$ ]{
\label{fig:exc_10000_4}
\psfrag{x}{$x$}
\psfrag{temp}{$u(x,t)$}
\includegraphics[width=0.45\textwidth,height=4cm]{FIGURES/posterior_slice_10000_ntpred=1000_zoom_x=0.25.eps}}
\caption{\em Comparison of predictive posterior estimates (posterior mean and $5\%$ and $95\%$ quantiles) with exact solution $u(x,t)$ at  $t=1.0$. These results were obtained with a prediction horizon (a) $T=0.05$ and (b) $T=0.1$.}
\label{fig:exc_posterior_3}
\end{figure}

Upon spatial discretization, we obtain a coupled system of ODEs:
\be
\label{eq:hed}
\dot{\bs{y}}_t+\bs{K}\bs{y}_t=\bs{0}
\ee
where $\bs{y}_t \in \RR^{1001}$ represents the solution at the nodes $x_i$, i.e.  $\bs{y}_t=\left[ u(x_1,t), u(x_2,t), \ldots,\right. $ $\left.\ldots, u(x_d,t)\right ]^T$. In contrast to existing approaches for the same diffusion equation (e.g.  \cite{Abdulle:2003,Abdulle:2007,Samaey:2006}) we do not exploit  mathematical properties  of the PDE in specifying the coarse-grained model but rely on data. This data is obtained upon  temporal discretization of \refeq{eq:hed}  where a time step $\delta t=0.0001$ was used. As a result at each time step we obtained a vector of observables $\bs{y}_t$ of dimension $d=1001$.
 This data was incorporated in the Bayesian model proposed using two hidden OU process ($M=2$) of dimension $K=2$ each. In particular we employed the online EM scheme previously discussed over blocks of length $L=10$ time steps and $N=100$ particles. In particular (see also Figure \ref{fig:pan}):
\bi
\item data over $20$ times steps $\delta t$, $\bs{y}_{1:20}$ (i.e. corresponding to total time $20 \delta t =0.002$) were ingested by the Bayesian reduced model, and
\item the latter was used to predict the evolution of the system over $500$ time steps (i.e.  total time $T=500 \delta t=0.05$).
\item The original solver of the governing PDE  was then re-initialized using the posterior mean estimate of the state of the system $\bs{y}_{520}$ and was run for further $20$ time steps. Using the additional data obtained $\bs{y}_{521:540}$, the Bayesian model was updated and the process described was repeated.
\ei
It is noted that the proposed Bayesian prediction  scheme results in a reduction of the number of fine scale integration time steps by a factor of $25$ ($T/20 \delta t=0.05/0.002$) leading to a significant acceleration of the simulation process. Figure \ref{fig:exc_posterior} depicts the posterior estimates of the solution at various time instants. In all cases these approximate very accurately the exact solution and these estimates improve as more are accumulated. One of the main advantages of the proposed approach is that apart from single-point estimates one can readily obtain credible intervals that quantify predictive uncertainties due to information loss by the use of the reduced-order dynamic model and the finite amount of data used to learn that model. As it is seen in Figure \ref{fig:exc_posterior} these envelop the exact solution and become tighter at larger times. 
As one would expect, when a larger predictive horizon $T=0.1$ (i.e. $1,000$ time steps $\delta t$) is used, as it can be seen in Figures \ref{fig:exc_posterior_2} and \ref{fig:exc_posterior_3}  the predictive uncertainty grows. Such a scheme however is twice as  efficient leading to a reduction of computational effort by a factor of $50$ (i.e. $T/20 \delta t=0.1/0.002$). Hence if the analyst is willing to tolerate the additional uncertainty, efficiency gains can be achieved. This supports the arguments made previously with regards to an {\em adaptive} Bayesian scheme where, the level of predictive uncertainty would be specified and the algorithm would automatically revert to the fine scale model in order to obtain more data and improve the predictive estimates.

\section{Conclusions}

The proposed modeling framework can extract interpretable reduced representations of  high-dimensional systems by employing  simple, low-dimensional processes. It simultaneously achieves  dimensionality reduction and learning of reduced dynamics.  The Bayesian framework adopted provides a generalization over   single-point  estimates obtained through maximum-likelihood procedures. It  can  quantify uncertainties associated with learning from finite amounts of data 
 and  produce probabilistic predictive estimates. The latter can be used to rigorously perform concurrent simulations with the microscopic model without the need of prescribing ad hoc upscaling and downscaling operators.

Critical to the efficacy of the proposed techniques is scalability particularly with regards to the large dimension $d$ of the original process. The  algorithms proposed  imply $O(d)$ order of operations.  Furthermore they can dynamically  {\em update } the coarse-grained models as more data become available.  In a typical scenario, the fine-scale model is reinitialized several times in order to obtain additional information about the system's evolution that is incorporated in the coarse-grained dynamics ``on the fly''.

The Bayesian, statistical perspective can readily be extended to the modeling  {\em stochastic} dynamical  systems. This would require generating more than one realizations of the original dynamics which can nevertheless be incorporated in the coarse-grained models using the same online EM scheme. In fact the loss of information that unavoidably takes place during the coarse-graining, results in probabilistic reduced-order  models even if the original model was deterministic. 
A critical question that offers opportunity for future research on the topic relates to structural learning and in  particularly with the  dimensionality of the representation, i.e. the number of hidden processes $\bs{x}_t^{(m)}$  needed  (denoted by $M$ in \refeq{eq:hidx}). Treating this as a model selection problem as it was done in the examples, assumes that there is a single, optimal  finite-dimensional representation.  Current research activities are centered around {\em nonparametric }  Bayesian priors over infinite combinatorial structures  based on the Dirichlet process paradigm and  infinite latent features models  (e.g. \cite{teh06hie,gri06inf,hel07non}). These offer an alternative  perspective by assuming that the number of building blocks  is potentially unbounded, and that the observables only manifest a sparse subset of those. As a result, the {\em cardinality}  of the coarse-grained model can be automatically determined from the data. 
Another aspect that warrants further investigation is prior modeling of the static parameters. Apart from the regularization effect this offers,  it can promote the discovery of desirable features, such as  slow-varying essential dynamics, sparse factors (e.g. $\bs{P}^{(m)}$ in \refeq{eq:obs}) which can advance the  interpretability of the results and facilitate the inference process. 





\newpage
\section*{Appendix}
This appendix discusses the sufficient statistics and update equations for the static parameters $\bs{\Theta}$ used in the probabilistic model proposed. In the first section we discuss parameters appearing in the reduced-order dynamics models  and in the second those appearing in the likelihood.

\subsection*{ Sufficient statistics for parameters appearing in the prior}
As discussed in section \ref{repres}, independent, isotropic OU processes are used as prior models for the latent, coarse-grained dynamics $\bs{x}_t^{(m)}\in \RR^K$ as well the process $\bs{z}_t \in \RR^M$ that models the frctional memberships to to each process $m$.
We therefore discuss the essential elements for the online EM computations  described in section \ref{sec:inference} (\cite{and03onl,and04par,and05onl}) for a general isotropic OU process in $\RR^n$ of the form:
\be
\label{eq:ougen}
d\bs{x}_t=-b (\bs{x}_t-\bs{q})dt+\bs{S}^{1/2} d\bs{W}_t
\ee
It is of interest to determine the parameters $\bs{\theta}=(b, \bs{q}, \bs{S})$. Let also $\pi(\bs{\theta})$ denote the prior on $\bs{\theta}$.
The readers can adjust the expressions below to any $\bs{x}_t^{(m)}$ or $\bs{z}_t$ since independent priors were used. 
Note that the stationary distribution of $\bs{x}_t$ is a Gaussian:
\be
\label{eq:oustat}
\nu_0(\bs{x})=\mathcal{N}(\bs{x} \mid \bs{q}, \bs{C}=\frac{1}{2b}\bs{S})
\ee
and the transition density $p(\bs{x}_t \mid \bs{x}_{t-1})$ assuming that equidistant time instants with time step $\delta t$ are considered, is given by:
\be
\label{eq:outr}
p(\bs{x}_t \mid \bs{x}_{t-1})=\mathcal{N}(\bs{x}_t \mid \bs{\mu}_{\delta t}(\bs{x}_{t-1}), \bs{S}_{\delta t} )
\ee
where:
\be
\bs{\mu}_{\delta t}(\bs{x}_{t-1})=\bs{x}_{t-1}-(1-e^{-b \delta t})(\bs{x}_{t-1}-\bs{q})
\ee
and:
\be
 \bs{S}_{\delta t}=\frac{1-e^{-2b \delta t} }{2 b } \bs{S}
\ee

Given a block of length $L$ with observables $\bs{y}_{1:L}$ and according to Equations (\ref{eq:sdlpost}) and (\ref{eq:batch}) we have that:
\be
\label{eq:ouss}
\begin{array}{ll}
\hat{Q}(\bs{\theta}^{(k-1)}, \bs{\theta})& =\left< -\frac{1}{2} \log \mid \bs{C} \mid -\frac{1}{2} (\bs{x}_1-\bs{q})^T \bs{C}^{-1} (\bs{x}_1-\bs{q}) \right. \nonumber  \\
& + \left. \sum_{t=2}^L  - \frac{1}{2} \log \mid \bs{S}_{\delta t} \mid -\frac{1}{2} (\bs{x}_t-\bs{q})^T \bs{C}^{-1} (\bs{x}_1-\bs{q}) \right> \nonumber \\
& +\log \pi(\bs{\theta})
\end{array}
\ee
where the brackets $<.>$ imply expectation with respect to $\hat{p}(\bs{x}_{1:L} \mid \bs{\theta}^{(k-1)}, \bs{y}_{1:L})$ as in \refeq{eq:batch}. In order to maximize $\hat{Q}(\bs{\Theta}^{(1:k-1)}, \bs{\Theta})$ as in \refeq{eq:mstepsa} one needs to solve the system of equations arising from $\frac{\pa \hat{Q}(\bs{\theta}^{(1:k-1)}, \bs{\theta}) }{ \pa \bs{\theta} }=\bs{0}$ 
These equations equations with respect to $\bs{\theta}$ are solved with fixed point iterations. They depend on the following $7$  sufficient statistics $\bs{\Phi}=\{\Phi_j\}_{j=1}^7$:
\be
\begin{array}{l}
 \Phi_1=<\bs{x}_1> \\
\Phi_2= < \bs{x}_1 \bs{x}^T_1 > \\
\Phi_3=\left< \sum_{t=2}^L\bs{x}_{t-1} \right> \\
\Phi_4= \left< \sum_{t=2}^L \bs{x}_t-\bs{x}_{t-1} \right> \\
\Phi_5= \left< \sum_{t=2}^L \bs{x}_{t-1} \bs{x}_{t-1}^T \right> \\
\Phi_6= \left< \sum_{t=2}^L (\bs{x}_t-\bs{x}_{t-1})\bs{x}_{t-1}^T \right> \\
\Phi_7= \left< \sum_{t=2}^L (\bs{x}_t-\bs{x}_{t-1})(\bs{x}_t-\bs{x}_{t-1})^T \right> 
\end{array}
\ee

\subsection*{ Sufficient statistics for parameters appearing in the likelihood}

The process a bit more involved in the case of the parameters appearing in the likelihood \refeq{eq:like} i.e. the projection matrices $\{ \bs{P}^{(m)} \}_{m=1}^M$ of dimension $d \times K$ and the covariance $\bs{\Sigma}$ which is a (positive definite) matrix of $d \times d$. In order to retain scalability in high-dimensional problems (i.e. $d>>1$) we assume a diagonal form of $\bs{\Sigma}=diag(\sigma_1^2, \sigma_2^2, \ldots, \sigma_d^2)$ which implies learning $d$ parameters rather than $d(d+1)/2$.

Denoting now by $\bs{\theta}=(\{ \bs{P}^{(m)} \}_{m=1}^M, \{\sigma_j^2\}_{j=1}^d )$ , $\pi(\bs{\theta})$ the prior and according to Equations (\ref{eq:sdlpost}) and (\ref{eq:batch}) we have that:
\be
\label{eq:likess}
\begin{array}{ll}
\hat{Q}(\bs{\theta}^{(k-1)}, \bs{\theta})& =\left<  \sum_{t=1}^L  - \frac{1}{2} \log \mid \bs{\Sigma} \mid -\frac{1}{2} (\bs{y}_t-\bs{W}_t \bs{X}_t )^T \bs{\Sigma}^{-1} (\bs{y}_t-\bs{W}_t \bs{X}_t) \right> \\
& + \log \pi(\bs{\theta})
\end{array}
\ee
Differentiation with respect to $ \bs{P}^{(m)} $ reveals that the stationary point must satisfy:
\be
\bs{A}^{(m)}=\sum_{n=1}^M \bs{P}^{(n)} \bs{B}^{(n,m)}
\ee
where  the sufficient statistics are:
\be
\label{eq:ap1}
\underbrace{\bs{A}^{(m)}}_{d \times K}=\left< \sum_{t=1}^L z_{t,m} \bs{y}_t (\bs{x}_t^{(m)})^T  \right>, \quad m=1,2,\ldots, M
\ee
and:
\be
\underbrace{ \bs{B}^{(n,m)} }_{K \times K} =\left< \sum_{t=1}^L z_{t,n} z_{t,m} \bs{x}_t^{(n)} (\bs{x}_t^{(m)})^T  \right>
\ee
In the absence of a prior $\pi(\bs{\theta})$ and if $\bs{P}_j^{(m)}$ and $\bs{A}^{(m)}_j$  represent the $j^{th}$ rows ($j=1,\ldots,d$) of the matrices $\bs{P}^{(m)}$ and $\bs{A}^{(m)}$ respectively, then \refeq{eq:ap1} implies:
\be
\begin{array}{ll}
\underbrace{ \left[ \begin{array}{llll} \bs{A}^{(1)}_j & \bs{A}^{(2)}_j & \ldots & \bs{A}^{(M)}_j \end{array} \right] }_{\bs{A}_j: (1 \times K~M)}= &
\underbrace{ \left[ \begin{array}{llll} \bs{P}^{(1)}_j &  \bs{P}^{(2)}_j & \ldots & \bs{P}^{(M)}_j \end{array} \right] }_{\bs{P}_j: (1\times K~M)} \\
& \underbrace{ \left[ \begin{array}{llll} \bs{B}^{(1,1)} &  \bs{B}^{(1,2)} & \ldots &  \bs{B}^{(1,M)} \\ \bs{B}^{(2,1)} & \bs{B}^{(2,2)} & \ldots &  \bs{B}^{(2,M)} \\ \ldots & \ldots & \ldots & \ldots \\ \bs{B}^{(M,1)} & \bs{B}^{(M,2)} & \ldots &  \bs{B}^{(M,M)} \end{array} \right] }_{\bs{B}: ( MK\times MK)}
\end{array}
\ee
This leads to  the following update equations for $\bs{P}_j^{(m)}, ~\forall j,m$:
\be
\bs{P}_j=\bs{A}_j \bs{B}^{-1}
\label{eq:ap2}
\ee
Note that the matrix $\bs{B}$ to be inverted is \underline{ independent } of the dimension of the observables $d$ ($d>>1$) and the inversion needs to be carried out once for all $j=1,\ldots,d$. Hence the {\em scaling of the update equations for $\bs{P}^{(m)}$ is $O(d)$} i.e. linear with respect to the dimensionality of the original system.

Furthermore, in the absence of a prior $\pi(\bs{\theta})$, differentiation with respect to $\sigma_j^{-2}$ ($j=1,\ldots,d$) leads to the following update equation:
\be
\begin{array}{ll}
L~\sigma_j^2 & =  \sum_{t=1}^L y_{t,j}^2-2 \bs{A}_j ~\bs{P}_j^T+\bs{P}_j~\bs{B} \bs{P}_j^T 
\end{array}
\ee
 In summary the sufficient statistics needed are the ones in Equations (\ref{eq:ap1}) and (\ref{eq:ap2}).

In the numerical examples in this paper a diffuse Gaussian prior was used for $\bs{P}^{(m)}$ with variance $100$ for each of the entries of the matrix. This leads to the addition of the term $1/100$ in the diagonal elements of the  $\bs{B}$ in \refeq{eq:ap2}. No priors were used for $\sigma_j^2$.

\newpage
\bibliographystyle{siam}
\bibliography{doe_early_career}

\begin{thebibliography}{100}

\bibitem{Abdulle:2007}
{\sc A.~Abdulle and B.~Engquist}, {\em Finite element heterogeneous multiscale
  methods with near optimal computational complexity}, MULTISCALE MODELING \&
  SIMULATION, 6 (2007), pp.~1059 -- 1084.

\bibitem{Abdulle:2003}
{\sc A.~Abdulle and E.~Weinan}, {\em Finite difference heterogeneous
  multi-scale method for homogenization problems}, JOURNAL OF COMPUTATIONAL
  PHYSICS, 191 (2003), pp.~18 -- 39.

\bibitem{Abraham:2002}
{\sc F.F. Abraham, R.~Walkup, H.J. Gao, M.~Duchaineau, T.D. De~la Rubia, and
  M.~Seager}, {\em Simulating materials failure by using up to one billion
  atoms and the world's fastest computer: Work-hardening}, PROCEEDINGS OF THE
  NATIONAL ACADEMY OF SCIENCES OF THE UNITED STATES OF AMERICA, 99 (2002),
  pp.~5783 -- 5787.

\bibitem{DBLP:conf/icde/Aggarwal09}
{\sc Charu~C. Aggarwal}, {\em A framework for clustering massive-domain data
  streams}, in ICDE, 2009, pp.~102--113.

\bibitem{agu00bay}
{\sc O.~Aguilar and M.~West}, {\em Bayesian dynamic factor models and variance
  matrix discounting for portfolio allocation}, Journal of Business and
  Economic Statistics, 18 (2000), p.~338–357.

\bibitem{and03onl}
{\sc C.~Andrieu and A.~Doucet}, {\em Online expectation-maximization type
  algorithms for parameter estimation in general state space models}, in IEEE
  International Conference on Acoustics, Speech, and Signal Processing, 6-10
  April 2003, pp.~69--72.

\bibitem{and04par}
{\sc C.~Andrieu, A.~Doucet, S.S. Singh, and V.B. Tadi\'c}, {\em Particle
  methods for change detection, identification and control}, in Proceedings of
  the IEEE, vol.~92, 2004, pp.~423--438.

\bibitem{and05onl}
{\sc C.~Andrieu, A.~Doucet, and V.B. Tadi\'c}, {\em Online simulation-based
  methods for parameter estimation in non linear non gaussian state-space
  models}, in Proc. IEEE CDC (invited paper), 2005.

\bibitem{AzranG06}
{\sc Arik Azran and Zoubin Ghahramani}, {\em Spectral methods for automatic
  multiscale data clustering}, in CVPR (1), 2006, pp.~190--197.

\bibitem{Bach:2006}
{\sc F.R. Bach and M.I. Jordan}, {\em Learning spectral clustering, with
  application to speech separation}, JOURNAL OF MACHINE LEARNING RESEARCH, 7
  (2006), pp.~1963 -- 2001.

\bibitem{DBLP:conf/dagm/BakirZT04}
{\sc G{\"o}khan~H. Bakir, Alexander Zien, and Koji Tsuda}, {\em Learning to
  find graph pre-images}, in DAGM-Symposium, 2004, pp.~253--261.

\bibitem{bea02inf}
{\sc M.~J. Beal, Z.~Ghahramani, and C.~E. Rasmussen}, {\em The infinite hidden
  markov model}, in Neural Information Processing Systems 14, T.G. Dietterich,
  S.~Becker, and Z.~Ghahramani, eds., MIT Press, 2002, pp.~577--585.

\bibitem{bis99lat}
{\sc C.~Bishop}, {\em Latent variable models}, in Learning in Graphical Models,
  M.~I. Jordan, ed., MIT Press, 1999, pp.~371--403.

\bibitem{ble03hie}
{\sc D~Blei, T~Griffiths, M~Jordan, and J~Tenenbaum}, {\em Hierarchical topic
  models and the nested chinese restaurant process}, in NIPS 2003, 2003.

\bibitem{ble06dyn}
{\sc D.~Blei and J.~Lafferty}, {\em Dynamic topic models}, in 23rd
  International Conference on Machine Learning, p.~2006.

\bibitem{ble03lat}
{\sc D.~Blei, A.~Ng, and M.~Jordan}, {\em Latent dirichlet allocation}, Journal
  of Machine Learning Research,  (2003), pp.~993--1022.

\bibitem{can07the}
{\sc E.~Cances, F.~Legoll, and G.~Stoltz}, {\em Theoretical and numerical
  comparison of some sampling methods for molecular dynamics}, Mathematical
  Modelling and Numerical Analysis, 41 (2007), p.~351.

\bibitem{cap01ten}
{\sc O.~Capp\'e}, {\em { Ten years of HMMs (An HMM bibliography)}}.
\newblock 2001.

\bibitem{DBLP:journals/corr/abs-0712-4273}
{\sc Olivier Capp{\'e} and Eric Moulines}, {\em Online em algorithm for latent
  data models}, CoRR, abs/0712.4273 (2007).

\bibitem{cap05inf}
{\sc O.~Capp\'e, E.~Moulines, and T.~Ryd\'en}, {\em {Inference in Hidden Markov
  Models}}, Springer-Verlag, 2005.

\bibitem{car07gen}
{\sc F.~Caron, M.~Davy, and A.~Doucet}, {\em {Generalized Polya urn for
  time-varying Dirichlet processes }}, in Proceedings of Uncertainty in
  Artificial Intelligence 2007, 2007.

\bibitem{cha08sca}
{\sc L.~Chac\'on}, {\em Scalable parallel implicit solvers for 3d
  magnetohydrodynamics}, Journal of Physics: Conference Series, 125 (2008),
  p.~012041.

\bibitem{DBLP:journals/tsmc/ChaerBG97}
{\sc Wassim~S. Chaer, Robert~H. Bishop, and Joydeep Ghosh}, {\em A
  mixture-of-experts framework for adaptive kalman filtering}, IEEE
  Transactions on Systems, Man, and Cybernetics, Part B, 27 (1997),
  pp.~452--464.

\bibitem{cha78sta}
{\sc C.~B. Chang and M.~Athans}, {\em State estimation for discrete systems
  with switching parameters}, IEEE Transactions on Aerospace and Electronic
  Systems, AES, 14, p.~418–424.

\bibitem{chi07fre}
{\sc C.~Chipot and A.~Pohorille}, eds., {\em Free energy calculations},
  Springer Series in Chemical Physics, 2007.

\bibitem{Chopin:2004}
{\sc N~Chopin}, {\em Central limit theorem for sequential monte carlo methods
  and its application to bayesian inference}, ANNALS OF STATISTICS, 32 (2004),
  pp.~2385 -- 2411.

\bibitem{Chorin:2000}
{\sc A.J. Chorin, O.H. Hald, and R.~Kupferman}, {\em {Optimal prediction and
  the Mori-Zwanzig representation of irreversible processes}}, PROCEEDINGS OF
  THE NATIONAL ACADEMY OF SCIENCES OF THE UNITED STATES OF AMERICA, 97 (2000),
  pp.~2968 -- 2973.

\bibitem{Chorin:2006}
\leavevmode\vrule height 2pt depth -1.6pt width 23pt, {\em Prediction from
  partial data, renormalization, and averaging}, JOURNAL OF SCIENTIFIC
  COMPUTING, 28 (2006), pp.~245 -- 261.

\bibitem{Chorin:2005}
{\sc A.J. Chorin and P.~Stinis}, {\em Problem reduction, renormalization, and
  memory}, Comm. Appl. Math. Comp. Sc., 1 (2005), pp.~1--27.

\bibitem{cic05non}
{\sc G.~Ciccotti, R.~Kapral, and A.~Sergi}, {\em Non-equilibrium molecular
  dynamics}, in Handbook of materials modeling, S.~Yip, ed., 2005,
  pp.~745--761.

\bibitem{Coifman:2008}
{\sc R.R. Coifman, I.G. Kevrekidis, S.~Lafon, M.~Maggioni, and B.~Nadler}, {\em
  Diffusion maps, reduction coordinates, and low dimensional representation of
  stochastic systems}, MULTISCALE MODELING \& SIMULATION, 7 (2008), pp.~842 --
  864.

\bibitem{Coifman:2005}
{\sc R.R. Coifman, S.~Lafon, A.B. Lee, M.~Maggioni, B.~Nadler, F.~Warner, and
  S.W. Zucker}, {\em Geometric diffusions as a tool for harmonic analysis and
  structure definition of data: Diffusion maps}, PROCEEDINGS OF THE NATIONAL
  ACADEMY OF SCIENCES OF THE UNITED STATES OF AMERICA, 102 (2005), pp.~7426 --
  7431.

\bibitem{DBLP:conf/sigmod/CranorJSS03}
{\sc Charles~D. Cranor, Theodore Johnson, Oliver Spatscheck, and Vladislav
  Shkapenyuk}, {\em Gigascope: A stream database for network applications}, in
  SIGMOD Conference, 2003, pp.~647--651.

\bibitem{dar09com}
{\sc E.~Darv\'e, J.~Solomon, and A.~Kiab}, {\em Computing generalized langevin
  equations and generalized fokker–planck equations}, PNAS, 106 (2009),
  pp.~10884--10889.

\bibitem{DBLP:conf/sac/SousaTTF06}
{\sc Elaine P.~M. de~Sousa, Agma J.~M. Traina, Caetano~Traina Jr., and Christos
  Faloutsos}, {\em Evaluating the intrinsic dimension of evolving data
  streams}, in SAC, 2006, pp.~643--648.

\bibitem{mor04fey}
{\sc P.~Del~Moral}, {\em Feynman-Kac Formulae: Genealogical and Interacting
  Particle Systems with Applications}, Springer New York, 2004.

\bibitem{del06seq}
{\sc P.~Del~Moral, A.~Doucet, and A.~Jasra}, {\em {Sequential Monte Carlo for
  Bayesian Computation (with discussion) }}, in Bayesian Statistics 8, Oxford
  University Press, 2006.

\bibitem{mor06seq}
{\sc P.~Del~Moral, A.~Doucet, and A.~Jasrau}, {\em {Sequential Monte Carlo
  Samplers}}, Journal of the Royal Statistical Society B, 68 (2006),
  pp.~411--436.

\bibitem{Dellago:1998}
{\sc C.~Dellago, P.G. Bolhuis, and D.~Chandler}, {\em Efficient transition path
  sampling: Application to lennard-jones cluster rearrangements}, JOURNAL OF
  CHEMICAL PHYSICS, 108 (1998), pp.~9236 -- 9245.

\bibitem{del06gra}
{\sc M.~Dellnitz, M.~Hessel-von Molo, P.~Metzner, R.~Preis, and C.~Sch\"utte},
  {\em Graph algorithms for dynamical systems}, in Analysis, Modeling and
  Simulation of Multiscale Problems, A.~Mielke, ed., Springer-Verlag,
  Heidelberg, 2006, p.~619–646.

\bibitem{dem77max}
{\sc A.P. Dempster, N.M. Laird, and D.B. Rubin}, {\em {Maximum likelihood from
  incomplete data via the EM algorithm (with discussion)}}, J. Roy. Statist.
  Soc. Ser. B, 39 (1977), pp.~1--38.

\bibitem{Deuflhard:2000}
{\sc P.~Deuflhard, W.~Huisinga, A.~Fischer, and C.~Sch\"utte}, {\em
  Identification of almost invariant aggregates in reversible nearly uncoupled
  markov chains}, LINEAR ALGEBRA AND ITS APPLICATIONS, 315 (2000), pp.~39 --
  59.

\bibitem{Deuflhard:2005}
{\sc P.~Deuflhard and M.~Weber}, {\em Robust perron cluster analysis in
  conformation dynamics}, LINEAR ALGEBRA AND ITS APPLICATIONS, 398 (2005),
  pp.~161 -- 184.

\bibitem{Donoho:2003}
{\sc D.L. Donoho and C.~Grimes}, {\em Hessian eigenmaps: Locally linear
  embedding techniques for high-dimensional data}, PROCEEDINGS OF THE NATIONAL
  ACADEMY OF SCIENCES OF THE UNITED STATES OF AMERICA, 100 (2003), pp.~5591 --
  5596.

\bibitem{dou01seq}
{\sc A.~Doucet, J.~F.~G. de~Freitas, and N.~J. Gordon}, eds., {\em {Sequential
  Monte Carlo Methods in Practice}}, Springer, New York, 2001.

\bibitem{dou00seq}
{\sc A.~Doucet, S.J. Godsill, and C.~Andrieu}, {\em {On Sequential Monte Carlo
  Sampling Methods for Bayesian Filtering}}, Statistics and Computing, 10,
  pp.~197--208.

\bibitem{e03het}
{\sc W.~E and B.~Engquist}, {\em The heterogeneous multi-scale methods}, Comm.
  Math. Sci., 1 (2003), p.~87.

\bibitem{e05ana}
{\sc W.~E, D.~Liu, and E.~Vanden-Eijnden}, {\em Analysis of multiscale methods
  for stochastic differential equations}, Comm. Pure Appl. Math., 58 (2005),
  pp.~1544--1585.

\bibitem{E:2005}
{\sc Weinan E, Weiqing Ren, and Eric Vanden-Eijnden}, {\em Finite temperature
  string method for the study of rare events.}, J Phys Chem B, 109 (2005),
  pp.~6688 -- 93.

\bibitem{Erban:2007}
{\sc R.~Erban, T.A. Frewen, X.A. Wang, T.C. Elston, R.~Coifman, B.~Nadler, and
  I.G. Kevrekidis}, {\em Variable-free exploration of stochastic models: A gene
  regulatory network example}, JOURNAL OF CHEMICAL PHYSICS, 126 (2007),
  p.~155103.

\bibitem{DBLP:conf/sigmod/FaloutsosKS07}
{\sc Christos Faloutsos, Tamara~G. Kolda, and Jimeng Sun}, {\em Mining large
  graphs and streams using matrix and tensor tools}, in SIGMOD Conference,
  2007, p.~1174.

\bibitem{fat04com}
{\sc I.~Fatkullin and E.~Vanden-Eijnden}, {\em A computational strategy for
  multiscale systems with applications to lorenz 96 model}, JOURNAL OF
  COMPUTATIONAL PHYSICS, 200 (2004), pp.~605--638.

\bibitem{lee07mul}
{\sc M.~Ferreira and H.~Lee}, {\em {Multiscale Modeling - A Bayesian
  Perspective}}, Springer Series in Statistics, Spinger, 2007.

\bibitem{mul06fer}
{\sc M.A.R. Ferreira, M.~West, H.~Lee, and D.~Higdon}, {\em Multiscale and
  hidden resolution time series models}, Bayesian Analysis, 2 (2006),
  pp.~294--314.

\bibitem{Fischer:2007}
{\sc A.~Fischer, S.~Waldhausen, I.~Horenko, E.~Meerbach, and C.~Sch\"utte},
  {\em Identification of biomolecular conformations from incomplete torsion
  angle observations by hidden markov models}, JOURNAL OF COMPUTATIONAL
  CHEMISTRY, 28 (2007), pp.~2453 -- 2464.

\bibitem{DBLP:conf/icml/FoxSJW08}
{\sc Emily~B. Fox, Erik~B. Sudderth, Michael~I. Jordan, and Alan~S. Willsky},
  {\em An hdp-hmm for systems with state persistence}, in ICML, 2008,
  pp.~312--319.

\bibitem{DBLP:conf/nips/FoxSJW08}
\leavevmode\vrule height 2pt depth -1.6pt width 23pt, {\em Nonparametric
  bayesian learning of switching linear dynamical systems}, in NIPS, 2008,
  pp.~457--464.

\bibitem{fra08hid}
{\sc C.~Franzke, D.~Crommelin, A.~Fischer, and A.~Majda1}, {\em A hidden markov
  model perspective on regimes and metastability in atmospheric flows}, J.
  Climate, 21 (2008), pp.~1740--1757.

\bibitem{bas08non}
{\sc B.~Ganapathysubramanian and N.~Zabaras}, {\em A non-linear dimension
  reduction methodology for generating data-driven stochastic input models},
  Journal of Computational Physics, 227 (2008), pp.~6612--6637.

\bibitem{Gear:2005}
{\sc C.W. Gear, T.J. Kaper, I.G. Kevrekidis, and A.~Zagaris}, {\em Projecting
  to a slow manifold: Singularly perturbed systems and legacy codes}, SIAM
  JOURNAL ON APPLIED DYNAMICAL SYSTEMS, 4 (2005), pp.~711 -- 732.

\bibitem{Gear:2003a}
{\sc C.W. Gear and I.G. Kevrekidis}, {\em Projective methods for stiff
  differential equations: Problems with gaps in their eigenvalue spectrum},
  SIAM JOURNAL ON SCIENTIFIC COMPUTING, 24 (2003), pp.~1091 -- 1106.

\bibitem{Gear:2003}
\leavevmode\vrule height 2pt depth -1.6pt width 23pt, {\em Telescopic
  projective methods for parabolic differential equations}, JOURNAL OF
  COMPUTATIONAL PHYSICS, 187 (2003), pp.~95 -- 109.

\bibitem{Gear:2002}
{\sc C.W. Gear, I.G. Kevrekidis, and C.~Theodoropoulos}, {\em {'Coarse'
  integration/bifurcation analysis via microscopic simulators: micro-Galerkin
  methods}}, COMPUTERS \& CHEMICAL ENGINEERING, 26 (2002), pp.~941 -- 963.

\bibitem{gha01int}
{\sc Z.~Ghahramani}, {\em {An Introduction to Hidden Markov Models and Bayesian
  Networks}}, Journal of Pattern Recognition and Artificial Intelligence, 15
  (2001), pp.~9--42.

\bibitem{gha04uns}
\leavevmode\vrule height 2pt depth -1.6pt width 23pt, {\em Unsupervised
  learning}, in Advanced Lectures on Machine Learning LNAI 3176, O.~Bousquet,
  G.~Raetsch, and U.~von Luxburg, eds., Springer-Verlag, 2004.

\bibitem{DBLP:journals/neco/GhahramaniH00}
{\sc Zoubin Ghahramani and Geoffrey~E. Hinton}, {\em Variational learning for
  switching state-space models}, Neural Computation, 12 (2000), pp.~831--864.

\bibitem{gha99fac}
{\sc Z.~Ghahramani and M.I. Jordan}, {\em {Factorial hidden Markov models}},
  Machine Learning, 29 (1999), pp.~245--273.

\bibitem{giv04ext}
{\sc D.~Givon, R.~Kupferman, and A.~Stuart}, {\em Extracting macroscopic
  dynamics: Model problems and algorithms}, Nonlinearity,  (2004).

\bibitem{Godsill:2004}
{\sc S.J. Godsill, A.~Doucet, and M.~West}, {\em Monte carlo smoothing for
  nonlinear time series}, JOURNAL OF THE AMERICAN STATISTICAL ASSOCIATION, 99
  (2004), pp.~156 -- 168.

\bibitem{gri06inf}
{\sc T.~Griffiths and Z.~Ghahramani}, {\em Infinite latent feature models and
  the indian buffet process}, in NIPS, 2006.

\bibitem{hai02geo}
{\sc E.~Hairer, C.~Lubich, and G.~Wanner}, {\em Geometric numerical
  integration}, Springer-Verlag, New York, 2002.

\bibitem{ham89new}
{\sc J.~D. Hamilton}, {\em A new approach to the economic analysis of
  nonstationary time series and the business cycle}, Econometrica, 57 (1989),
  p.~357–384.

\bibitem{ham08tow}
{\sc G.~Hammond, P.~Lichtner, and C.~Lu}, {\em {Toward petascale computing in
  geosciences: application to the Hanford 300 Area}u}, Journal of Physics:
  Conference Series, 125 (2008), p.~012051.

\bibitem{har76bay}
{\sc P.~J. Harrison and C.~F. Stevens}, {\em Bayesian forecasting (with
  discussion)}, Royal Statistical Society B,  (1976).

\bibitem{hel07non}
{\sc K.A. Heller and Z.~Ghahramani}, {\em {A Nonparametric Bayesian Approach to
  Modeling Overlapping Clusters}}, in 11th International Conference on AI and
  Statistics (AISTATS 2007), 2007.

\bibitem{hel08sta}
{\sc K.A. Heller, S~Williamson, and Z.~Ghahramani}, {\em Statistical models for
  partial membership}, in Proceedings of the 25th International Conference on
  Machine Learning, 2008.

\bibitem{hin02tra}
{\sc G.~Hinton}, {\em Training products of experts by minimizing contrastive
  divergence}, Neural Computation, 14 (2002).

\bibitem{hoo92non}
{\sc W.G. Hoover}, {\em Nonequilibrium molecular dynamics}, Nuclear Physics,
  (1992), pp.~523--536.

\bibitem{Horenko:2008}
{\sc I.~Horenko}, {\em On simultaneous data-based dimension reduction and
  hidden phase identification}, JOURNAL OF THE ATMOSPHERIC SCIENCES, 65 (2008),
  pp.~1941 -- 1954.

\bibitem{Horenko:2007}
{\sc I.~Horenko, R.~Klein, S.~Dolaptchiev, and C.~Schuette}, {\em Automated
  generation of reduced stochastic weather models i: Simultaneous dimension and
  model reduction for time series analysis}, MULTISCALE MODELING \& SIMULATION,
  6 (2007), pp.~1125 -- 1145.

\bibitem{hor07dat}
{\sc I.~Horenko, F.~Noe, C.~Hartmann, and C.~Sch\"utte}, {\em Data-based
  parameter estimation of generalized multidimensional langevin processes},
  Physical Review E, 78 (2007), p.~016706.

\bibitem{Horenko:2006}
{\sc I.~Horenko, J.~Schmidt-Ehrenberg, and C.~Sch\"utte}, {\em Set-oriented
  dimension reduction: Localizing principal component analysis via hidden
  markov models}, COMPUTATIONAL LIFE SCIENCES II, PROCEEDINGS, 4216 (2006),
  pp.~74 -- 85.

\bibitem{Horenko:2008b}
{\sc I.~Horenko and C.~Sch\"utte}, {\em Likelihood-based estimation of
  multidimensional langevin models and its application in biomolecular
  dynamics}, MULTISCALE MODELING \& SIMULATION, 7 (2008), pp.~731 -- 773.

\bibitem{hot33ana}
{\sc H.~Hotelling}, {\em Analysis of a complex of statistical variables into
  principal components}, Journal of Educational Psychology, 24 (1933),
  pp.~417--441.

\bibitem{DBLP:conf/vldb/KoudasIM00}
{\sc Piotr Indyk, Nick Koudas, and S.~Muthukrishnan}, {\em Identifying
  representative trends in massive time series data sets using sketches}, in
  VLDB 2000, Proceedings of 26th International Conference on Very Large Data
  Bases, September 10-14, 2000, Cairo, Egypt, Amr~El Abbadi, Michael~L. Brodie,
  Sharma Chakravarthy, Umeshwar Dayal, Nabil Kamel, Gunter Schlageter, and
  Kyu-Young Whang, eds., Morgan Kaufmann, 2000, pp.~363--372.

\bibitem{jac91ada}
{\sc R.~A. Jacobs, M.~I. Jordan, and G.~E. Nowlan, S. J. nad~Hinton}, {\em
  Adaptive mixtures of local experts}, Neural Computation, 3 (1991),
  p.~79–87.

\bibitem{jac94hie}
{\sc M.~I. Jordan and R.~A. Jacobs}, {\em { Hierarchical mixtures of experts
  and the EM algorithm}}, Neural Computation, 6 (1994), p.~181–214.

\bibitem{kal02atm}
{\sc E.~Kalnay}, {\em Atmospheric Modeling, Data Assimilation and
  Predictability}, Cambridge University Press, 2002.

\bibitem{DBLP:journals/cviu/KanthAAS99}
{\sc Kothuri Venkata~Ravi Kanth, Divyakant Agrawal, Amr~El Abbadi, and Ambuj~K.
  Singh}, {\em Dimensionality reduction for similarity searching in dynamic
  databases}, Computer Vision and Image Understanding, 75 (1999), pp.~59--72.

\bibitem{Kavousanakis:2007}
{\sc M.E. Kavousanakis, R.~Erban, A.G. Boudouvis, C.W. Gear, and I.G.
  Kevrekidis}, {\em Projective and coarse projective integration for problems
  with continuous symmetries}, JOURNAL OF COMPUTATIONAL PHYSICS, 225 (2007),
  pp.~382 -- 407.

\bibitem{DBLP:conf/sigmod/KeoghCMP01}
{\sc Eamonn~J. Keogh, Kaushik Chakrabarti, Sharad Mehrotra, and Michael~J.
  Pazzani}, {\em Locally adaptive dimensionality reduction for indexing large
  time series databases}, in SIGMOD Conference, 2001, pp.~151--162.

\bibitem{Kevrekidis:2004}
{\sc I.G. Kevrekidis, C.W. Gear, and G.~Hummer}, {\em Equation-free: The
  computer-aided analysis of complex multiscale systems}, AICHE JOURNAL, 50
  (2004), pp.~1346 -- 1355.

\bibitem{kev03equ}
{\sc IG~Kevrekidis, CW~Gear, JM~Hyman, PG~Kevrekidis, O~Runborg, and
  K~Theodoropoulos}, {\em Equation-free multiscale computation: enabling
  microscopic simulators to perform system-level tasks}, Communications in
  Mathematical Sciences, 1 (2003), pp.~715--762.

\bibitem{DBLP:conf/sigmod/KornJF97}
{\sc Flip Korn, H.~V. Jagadish, and Christos Faloutsos}, {\em Efficiently
  supporting ad hoc queries in large datasets of time sequences}, in SIGMOD
  Conference, 1997, pp.~289--300.

\bibitem{kou07sto}
{\sc P.S. Koutsourelakis}, {\em Stochastic upscaling in solid mechanics: An
  excercise in machine learning}, Journal of Computational Physics, 226 (2007),
  pp.~301--325.

\bibitem{Lafon:2006}
{\sc S.~Lafon and A.B. Lee}, {\em Diffusion maps and coarse-graining: A unified
  framework for dimensionality reduction, graph partitioning, and data set
  parameterization}, IEEE TRANSACTIONS ON PATTERN ANALYSIS AND MACHINE
  INTELLIGENCE, 28 (2006), pp.~1393 -- 1403.

\bibitem{Laio:2002}
{\sc A.~Laio and M.~Parrinello}, {\em Escaping free-energy minima}, PROCEEDINGS
  OF THE NATIONAL ACADEMY OF SCIENCES OF THE UNITED STATES OF AMERICA, 99
  (2002), pp.~12562 -- 12566.

\bibitem{Lam:1993}
{\sc S.H. Lam}, {\em Using csp to understand complex chemical-kinetics},
  COMBUSTION SCIENCE AND TECHNOLOGY, 89 (1993), pp.~375 -- 404.

\bibitem{Lam:1994}
{\sc S.H. Lam and D.A. Goussis}, {\em The csp method for simplifying kinetics},
  INTERNATIONAL JOURNAL OF CHEMICAL KINETICS, 26 (1994), pp.~461 -- 486.

\bibitem{lei04sim}
{\sc B.~Leimkuhler and S.~Reich}, {\em Simulating Hamiltonian Dynamics},
  Cambridge University Press, 2004.

\bibitem{li08eff}
{\sc T.~Li, A.~Abdulle, and W.~E}, {\em Effectiveness of implicit methods for
  stiff stochastic differential equations}, Comm. Comput. Phys., 3 (2008),
  pp.~295--307.

\bibitem{lig08ter}
{\sc D.O. Lignell, J.H. Chen, and E.S. Richardson}, {\em Terascale direct
  numerical simulations of turbulent combustion — fundamental understanding
  towards predictive models}, Journal of Physics: Conference Series, 125
  (2008), p.~012031.

\bibitem{liu09dyn}
{\sc M.~Liu, F.~West}, {\em {A dynamic modelling strategy for Bayesian computer
  model emulation}}, Bayesian Analysis, 4 (2009), pp.~393--412.

\bibitem{Maas:1992}
{\sc U.~Maas and S.B. Pope}, {\em Simplifying chemical-kinetics - intrinsic
  low-dimensional manifolds in composition space}, COMBUSTION AND FLAME, 88
  (1992), pp.~239 -- 264.

\bibitem{mez05spe}
{\sc I.~Mezi\'c}, {\em Spectral properties of dynamical systems, model
  reduction and decompositions}, Nonlinear Dynamics, 41, pp.~309--325.

\bibitem{mil99dat}
{\sc R.N. Miller, E.F. Carter, and S.T. Blue}, {\em Data assimilation into
  nonlinear stochastic models}, Tellus, 51A (1999), pp.~167--194.

\bibitem{Nadler:2006}
{\sc B.~Nadler, S.~Lafon, R.R. Coifman, and I.G. Kevrekidis}, {\em Diffusion
  maps, spectral clustering and reaction coordinates of dynamical systems},
  APPLIED AND COMPUTATIONAL HARMONIC ANALYSIS, 21 (2006), pp.~113 -- 127.

\bibitem{ott04loc}
{\sc E.~Ott, B.R. Hunt, I.~Szunyogh, A.V. Zimin, E.J. Kostelich, M.~Corazza,
  E.~Kalnay, D.~Patil, and J.A. Yorke}, {\em {A local ensemble Kalman filter
  for atmospheric data assimilation}}, Tellus A, 56 (2004), pp.~415--428.

\bibitem{DBLP:journals/vldb/PapadimitriouBF04}
{\sc Spiros Papadimitriou, Anthony Brockwell, and Christos Faloutsos}, {\em
  Adaptive, unsupervised stream mining}, VLDB J., 13 (2004), pp.~222--239.

\bibitem{DBLP:conf/vldb/PapadimitriouSF05}
{\sc Spiros Papadimitriou, Jimeng Sun, and Christos Faloutsos}, {\em Streaming
  pattern discovery in multiple time-series}, in VLDB, 2005, pp.~697--708.

\bibitem{pea01lin}
{\sc K.~Pearson}, {\em On lines and planes of closest fit to systems of points
  in space}, The London, Edinburgh and Dublin Philosophical Magazine and
  Journal of Science, 2 (1901), pp.~, 559–572.

\bibitem{pet97num}
{\sc L.R. Petzold, L.O. Jay, and J.~Yen}, {\em Numerical solution of highly
  oscillatory ordinary differential equations}, Acta Numerica, 7 (1997),
  pp.~437--483.

\bibitem{ren07red}
{\sc Z.~Ren and S.B. Pope}, {\em Reduced description of complex dynamics in
  reactive systems}, Journal of Physical Chemistry A, 111 (2007),
  pp.~8464--8474.

\bibitem{RicoMartinez:2004}
{\sc R.~Rico-Martinez, C.W. Gear, and I.G. Kevrekidis}, {\em Coarse projective
  kmc integration: forward/reverse initial and boundary value problems},
  JOURNAL OF COMPUTATIONAL PHYSICS, 196 (2004), pp.~474 -- 489.

\bibitem{Roweis:2000}
{\sc S.T. Roweis and L.K. Saul}, {\em Nonlinear dimensionality reduction by
  locally linear embedding}, 290 (2000), p.~2323.

\bibitem{ryd94con}
{\sc T.~Ryd\'en}, {\em Consistent and asymptotically normal parameter estimates
  for hidden markov models}, Ann. Stat., 22 (1994), p.~1884–1895.

\bibitem{DBLP:conf/sigmod/SakuraiPF05}
{\sc Yasushi Sakurai, Spiros Papadimitriou, and Christos Faloutsos}, {\em
  Braid: Stream mining through group lag correlations}, in SIGMOD Conference,
  2005, pp.~599--610.

\bibitem{Samaey:2006}
{\sc G.~Samaey, I.G. Kevrekidis, and D.~Roose}, {\em Patch dynamics with
  buffers for homogenization problems}, JOURNAL OF COMPUTATIONAL PHYSICS, 213
  (2006), pp.~264 -- 287.

\bibitem{sat00onl}
{\sc M.~Sato and S.~Ishii}, {\em On-line em algorithm for the normalized
  gaussian network}, Neural Computation, 12 (2000), pp.~407--432.

\bibitem{saw06mod}
{\sc Acharya A. (2006) 195 6287-6311. Sawant, A.}, {\em Model reduction via
  parametrized locally invariant manifolds: Some examples}, Computer Methods in
  Applied Mechanics and Engineering, 195 (2006), pp.~6287--6311.

\bibitem{sch97ker}
{\sc B.~Sch\"olkopf, A.~J. Smola, and K.-R. Mueller}, {\em Kernel principal
  component analysis}, in LECTURE NOTES IN COMPUTER SCIENCE, SPRINGER VERLAG
  KG, 1997, pp.~583--588.

\bibitem{sch07fas}
{\sc S.~Vishwanathan S.~V.N. Schraudolph, N. N.~Gunter}, {\em Fast iterative
  kernel pca}, in ADVANCES IN NEURAL INFORMATION PROCESSING SYSTEMS, 2007,
  pp.~1225--1232.

\bibitem{Shi:2000}
{\sc J.B. Shi and J.~Malik}, {\em Normalized cuts and image segmentation}, IEEE
  TRANSACTIONS ON PATTERN ANALYSIS AND MACHINE INTELLIGENCE, 22 (2000), pp.~888
  -- 905.

\bibitem{shu91dyn}
{\sc R.~H. Shumway and D.~S. Stoffer}, {\em Dynamic linear models with
  switching.}, J. Amer. Stat. Assoc., 86 (1991), p.~763–769.

\bibitem{sre05tim}
{\sc N.~Srebro and S.~Roweis}, {\em Time-varying topic models using dependent
  dirichlet processes}, Tech. Report UTML TR 2005–003,, Department of
  Computer Science, University of Toronto, 2005.

\bibitem{DBLP:conf/pakdd/SunPF06}
{\sc Jimeng Sun, Spiros Papadimitriou, and Christos Faloutsos}, {\em
  Distributed pattern discovery in multiple streams}, in PAKDD, 2006,
  pp.~713--718.

\bibitem{tao09non}
{\sc M.~Tao, H.~Owhadi, and J.E. Marsden}, {\em {Non-intrusive and structure
  preserving multiscale integration of stiff ODEs, SDEs and Hamiltonian systems
  with hidden slow dynamics via flow averaging}}.
\newblock http://arxiv.org/abs/0908.1241.

\bibitem{tay08pet}
{\sc M.A. Taylor, J.~Edwards, and A.~St.Cyr}, {\em Petascale atmospheric models
  for the communit climate system model: new developments and evaluation of
  scalable dynamical cores}, Journal of Physics: Conference Series, 125 (2008),
  p.~012023.

\bibitem{teh06hie}
{\sc Y~Teh, M~Jordan, M~Beal, and D~Blei}, {\em Hierarchical dirichlet
  processes}, Journal of the American Statistical Association,  (2006).

\bibitem{Tenenbaum:2000}
{\sc J.B. Tenenbaum, V.~de~Silva, and J.C. Langford}, {\em A global geometric
  framework for nonlinear dimensionality reduction}, SCIENCE, 290 (2000),
  p.~2319.

\bibitem{DBLP:conf/vldb/TengCY03}
{\sc Wei-Guang Teng, Ming-Syan Chen, and Philip~S. Yu}, {\em A regression-based
  temporal pattern mining scheme for data streams}, in VLDB, 2003, pp.~93--104.

\bibitem{tuc92rev}
{\sc M.~Tuckerman, B.~J. Berne, and G.J. Martyna}, {\em Reversible multiple
  time scale molecular dynamics}, J. Chem. Phys., 97 (1992), p.~1990–2001.

\bibitem{vas08par}
{\sc N.~Vaswani}, {\em Particle filtering for large dimensional state spaces
  with multimodal observation likelihoods}, IEEE Trans. Signal Processing,
  (2008), pp.~4583--4597.

\bibitem{Voter:2002}
{\sc A.F. Voter, F.~Montalenti, and T.C. Germann}, {\em Extending the time
  scale in atomistic simulation of materials}, ANNUAL REVIEW OF MATERIALS
  RESEARCH, 32 (2002), pp.~321 -- 346.

\bibitem{Voter:2002a}
{\sc A.F. Voter, F.~Montalenti, T.C. Germann, B.P. Uberuaga, and J.A. Sprague},
  {\em Accelerated molecular dynamics methods.}, ABSTRACTS OF PAPERS OF THE
  AMERICAN CHEMICAL SOCIETY, 223 (2002), pp.~238--COMP.

\bibitem{wan08con}
{\sc C.~Wang, D.~Blei, and D.~Heckerman}, {\em Continuous time dynamic topic
  models}, in Uncertainty in Artificial Intelligence [UAI], p.~2008.

\bibitem{war06onl}
{\sc M.~K. Warmuth and D.~Kuzmin}, {\em On-line variance minimization}, in
  Proceedings of the 19th Annual Conference on Learning Theory (COLT 06),
  Carnegie Mellon, Pittsburg PA, 2006.

\bibitem{wea09var}
{\sc J.~Weare, D.~Givon, and P.~Stinis}, {\em Variance reduction for particle
  filters of systems with time-scale separation}, IEEE Trans. Signal Proc., 57
  (2009), p.~424.

\bibitem{wei93tim}
{\sc A.S. Weigend and N.A. Gershenfeld}, eds., {\em Time Series Prediction:
  Forecasting The Future And Understanding The Past}, Santa Fe Institute
  Studies in the Sciences of Complexity, Westview Press, 1993.

\bibitem{wes97bay}
{\sc M.~West and P.~Harrison}, {\em Bayesian Forecasting and Dynamic Models},
  New York: Springer-Verlag, 1997.

\bibitem{wik98hie}
{\sc C.K. Wikle, L.M. Berliner, and N.~Cressie}, {\em {Hierarchical Bayesian
  space-time models}}, Environmental and Ecological Statistics, 5 (1998),
  pp.~117--154.

\bibitem{DBLP:conf/cidr/YaoG03}
{\sc Yong Yao and Johannes Gehrke}, {\em Query processing in sensor networks},
  in CIDR, 2003.

\bibitem{DBLP:conf/icde/YiSJJFB00}
{\sc Byoung-Kee Yi, Nikolaos Sidiropoulos, Theodore Johnson, H.~V. Jagadish,
  Christos Faloutsos, and Alexandros Biliris}, {\em Online data mining for
  co-evolving time sequences}, in ICDE, 2000, pp.~13--22.

\bibitem{Zagaris:2004}
{\sc A.~Zagaris, H.G. Kaper, and T.J. Kaper}, {\em Analysis of the
  computational singular perturbation reduction method for chemical kinetics},
  JOURNAL OF NONLINEAR SCIENCE, 14 (2004), pp.~59 -- 91.

\bibitem{DBLP:conf/vldb/ZhuS02}
{\sc Yunyue Zhu and Dennis Shasha}, {\em Statstream: Statistical monitoring of
  thousands of data streams in real time}, in VLDB, 2002, pp.~358--369.

\bibitem{zwa01non}
{\sc R.~Zwanzig}, {\em Noequilibrium Statistical Mechanics}, Oxford University
  Press, New York, 2001.

\end{thebibliography}

\end{document}